%% file: arxiv.tex
\documentclass{article}

\usepackage[final,nonatbib]{neurips_2022}
\usepackage[utf8]{inputenc} 
\usepackage[T1]{fontenc}    
\usepackage{hyperref}       
\usepackage{url}            
\usepackage{booktabs}       
\usepackage{amsfonts}       
\usepackage{nicefrac}       
\usepackage{microtype}      
\usepackage[dvipsnames]{xcolor}         
\usepackage{tikz}
\usepackage{pgfplots}
\usepackage{textcomp}

\usepackage[bottom]{footmisc}
\usepackage{times}
\usepackage{floatrow}
\usepackage[tableposition=top]{caption}
\usepackage{url}
\usepackage[utf8]{inputenc}
\usepackage{amsmath,bm, amssymb} 
\usepackage{multirow}
\usepackage{stmaryrd}
\usepackage{booktabs}
\usepackage{subcaption}
\usepackage{xspace}
\usepackage{array}
\usepackage[numbers]{natbib}
\usepackage{mathtools}
\usepackage{soul}

\title{CroCo: Self-Supervised Pre-training \\ for 3D Vision Tasks by Cross-View Completion}
\author{Philippe Weinzaepfel \And Vincent Leroy \And Thomas Lucas \AND Romain Br\'egier \And Yohann Cabon \And Vaibhav Arora \And Leonid Antsfeld \AND Boris Chidlovskii \And Gabriela Csurka \And J\'er\^ome Revaud \AND \textmd{NAVER LABS Europe} \\ 
{\small \url{https://europe.naverlabs.com/research/computer-vision/croco/} }
}

\usepackage{xspace}
\input{plots/plotabbrev}
\makeatletter
\DeclareRobustCommand\onedot{\futurelet\@let@token\@onedot}
\def\@onedot{\ifx\@let@token.\else.\null\fi\xspace}
\def\eg{\emph{e.g}\onedot} 
\def\ie{\emph{i.e}\onedot} 
 
\def\etc{\emph{etc}\onedot}
\def\vs{\emph{vs}\onedot}
\def\wrt{\emph{w.r.t}\onedot}

\makeatother

\usepackage{pifont}
\newcommand{\cmark}{\ding{51}}
\newcommand{\xmark}{\ding{55}}

\newcommand{\PARp}[1]{\noindent{\bf{#1}}}
\newcommand{\PAR}[1]{\PARp{#1}.}
\newcommand{\rsp}{\vspace{-0.25cm} }

\begin{document}

\maketitle

\vspace{-0.35cm}

\begin{abstract}

\vspace{-0.05cm}

Masked Image Modeling (MIM) has 
recently been established as a potent pre-training paradigm. 
A pretext task is constructed by masking patches in an input image, and this masked content is then predicted by a neural network using visible patches as sole input. This pre-training leads to state-of-the-art performance when finetuned for high-level semantic tasks, \eg image classification and object detection.
In this paper we instead seek to learn representations that transfer well to a wide variety of 3D vision and lower-level geometric downstream tasks, such as depth prediction or optical flow estimation.
Inspired by MIM, we propose an unsupervised representation learning task trained from \emph{pairs} of images showing the same scene from different viewpoints.
More precisely, we propose the pretext task of {\it cross-view completion} where the first input image is partially masked, and this masked content has to be reconstructed 
from the visible content and the second image.
In single-view MIM, the masked content often cannot be inferred precisely from the visible portion only, so the model learns to act as a prior influenced by high-level semantics.
In contrast, this ambiguity can be resolved with cross-view completion 
from the second unmasked image, on the condition that the model is able to understand the spatial relationship between the two images.
Our experiments show that our pretext task leads to significantly improved performance for monocular 3D vision downstream tasks such as depth estimation. 
In addition, our model can be 
directly applied to binocular downstream tasks like optical flow or relative camera pose estimation, 
for which we obtain competitive results without bells and whistles, \ie, using a generic architecture without any task-specific design.

\vspace{-0.2cm}
\end{abstract}

\begin{figure}[h]
\centering
\includegraphics[width=\linewidth]{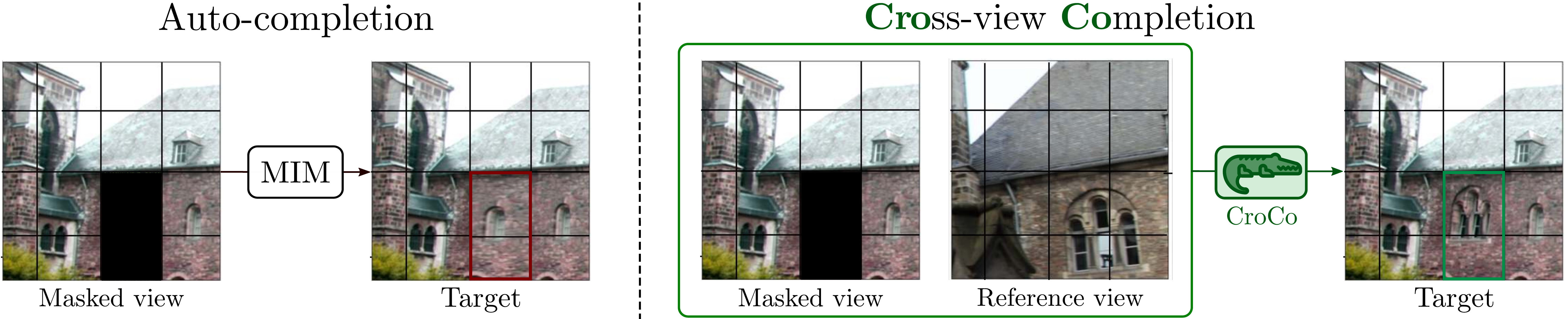} \\[-0.3cm]
\caption{\textbf{Auto-completion \vs Cross-view completion tasks.} 
    \textit{Left:} given a masked image, a model
trained for auto-completion can only leverage the visible context to fill-in the blanks and thus relies mainly on high-level semantic information.
    \textit{Right:} given an additional view of the same scene, cross-view completion makes precise reconstruction possible, 
    assuming that the model is able to understand both the scene geometry and the spatial relationship between the two images.
    }
\label{fig:task}
\end{figure}

\section{Introduction}

Self-supervised learning for model pre-training has allowed to achieve state-of-the-art performance in a number of high-level computer vision tasks, such as image classification or object detection. Contrary to traditional supervised learning, these models enable the use of 
unlabelled data via carefully designed pretext tasks.
A popular method for self-supervision is instance discrimination~\cite{caron2018deep,CaronNIPS20SwAV,ChenICML20SimCLR,DosovitskiyNIPS14DiscriminativeUnsupervisedFeatureLearningCNN,grill2020byol,HeCVPR20MoCo,wu2018npid} which constructs a pretext task by learning representations that are invariant to various data augmentations. 
More recently, Masked Image Modeling (MIM)~\cite{BaoICLR22BEiT,ChenICML20GenerativepretrainingFromPixels,Chen2022context,Fang22corrupted,he2021mae,PathakCVPR16ContextEncoders,Zhao2021selfsupervised} 
has emerged as a powerful alternative for self-supervision. Inspired by BERT~\cite{DevlinNAACL19BERT}, 
these models are trained using an auto-completion pretext task. An encoder, usually a Vision Transformer (ViT)~\cite{DosovitskiyICLR21ViT}, takes a partial view of an image input, obtained by splitting the image into patches and masking some of them, and encodes it into a latent representation. 
The masked patches are then predicted by a decoder using the latent representation.
To solve the task, the model must leverage the context given by the visible portion and act as a prior for the ambiguous content that cannot be deduced from them.
In practice these models are typically trained on object-centric datasets such as ImageNet~\cite{ILSVRC15imagenet} and thus tend to learn high-level semantic information; that makes them well suited for tasks such as image classification or object detection~\cite{bachmann2022multimae,he2021mae,li2022exploring}.

\begin{figure}[t]
    \centering
    \includegraphics[width=2.2cm,height=2.2cm]{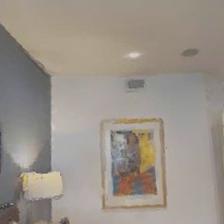} \hfill
    \includegraphics[width=2.2cm,height=2.2cm]{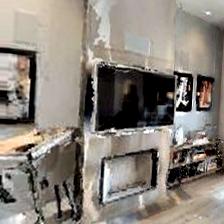} \hfill
    \includegraphics[width=2.2cm,height=2.2cm]{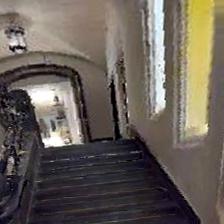} \hfill
    \includegraphics[width=2.2cm,height=2.2cm]{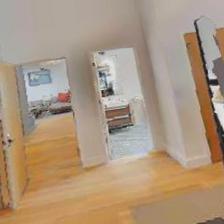} \hfill
    \includegraphics[width=2.2cm,height=2.2cm]{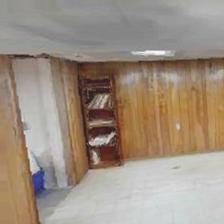} \hfill
    \includegraphics[width=2.2cm,height=2.2cm]{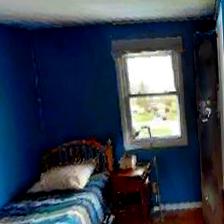} \hfill    
    \\

    \includegraphics[width=2.2cm,height=2.2cm]{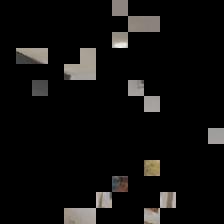} \hfill
    \includegraphics[width=2.2cm,height=2.2cm]{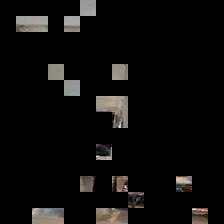} \hfill
    \includegraphics[width=2.2cm,height=2.2cm]{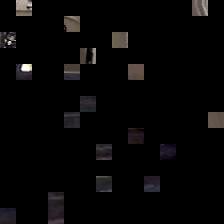} \hfill
    \includegraphics[width=2.2cm,height=2.2cm]{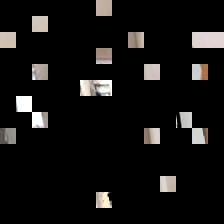} \hfill
    \includegraphics[width=2.2cm,height=2.2cm]{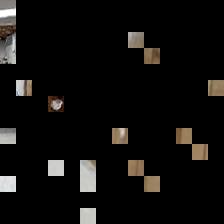} \hfill
    \includegraphics[width=2.2cm,height=2.2cm]{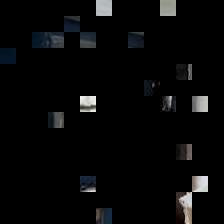} \hfill    
    \\
    
    \includegraphics[width=2.2cm,height=2.2cm]{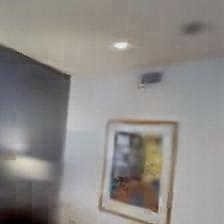} \hfill
    \includegraphics[width=2.2cm,height=2.2cm]{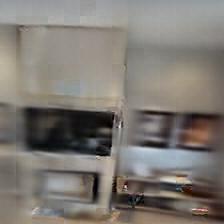} \hfill
    \includegraphics[width=2.2cm,height=2.2cm]{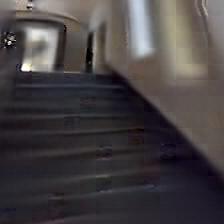} \hfill
    \includegraphics[width=2.2cm,height=2.2cm]{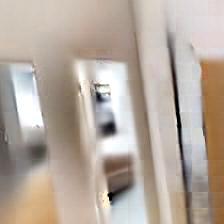} \hfill    
    \includegraphics[width=2.2cm,height=2.2cm]{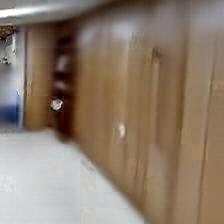} \hfill
    \includegraphics[width=2.2cm,height=2.2cm]{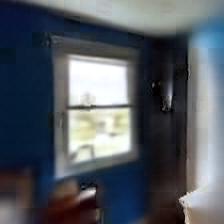} \hfill    
    \\
     
    \includegraphics[width=2.2cm,height=2.2cm]{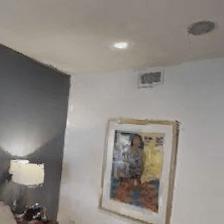} \hfill
    \includegraphics[width=2.2cm,height=2.2cm]{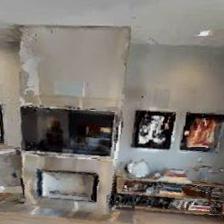} \hfill
    \includegraphics[width=2.2cm,height=2.2cm]{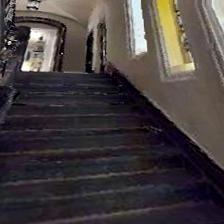} \hfill
    \includegraphics[width=2.2cm,height=2.2cm]{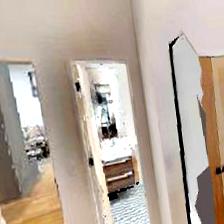} \hfill
    \includegraphics[width=2.2cm,height=2.2cm]{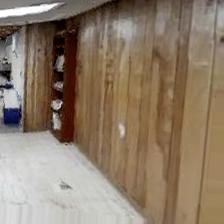} \hfill
    \includegraphics[width=2.2cm,height=2.2cm]{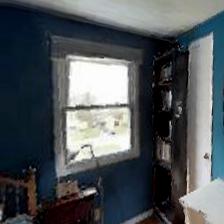} \hfill    
    \\
    \caption{\textbf{Reconstruction examples from CroCo} on scenes unseen during training. From top to bottom: reference image (input), masked image (input), CroCo output, original target image. }
    \rsp{}
    \label{fig:reconstructions}
\end{figure}

In this paper, we propose a self-supervised training objective specially designed to learn 
3D geometry from unlabeled data, named \textit{Cross-view Completion}, or \textit{CroCo} in short.
Given two images depicting the same scene, random parts of the first input image are masked and then predicted by the model using both (1) the visible parts of this first image, as well as (2) a second image called \emph{reference} image as it depicts the same scene from a different point of view, see Figure~\ref{fig:task}.
While multi-view image completion has a long history in image editing~\cite{CriminisiTIP04RegionFillingObjectRemovalExemplarBasedImageInpainting,Sameh2014ImageCompletion}, we are the first to explore its potential as a self-supervised representation learning tool.
In contrast to single-view completion, the proposed pretext task of cross-view completion allows to perform masked image modeling conditionally on a second  view.
In this case, most of the ambiguity can be resolved by reasoning about the 
scene geometry and spatial relationship between the two views.
This is illustrated in the reconstruction examples of our model in Figure~\ref{fig:reconstructions}.

\begin{figure}[ttt]
\centering
\includegraphics[width=0.9\linewidth]{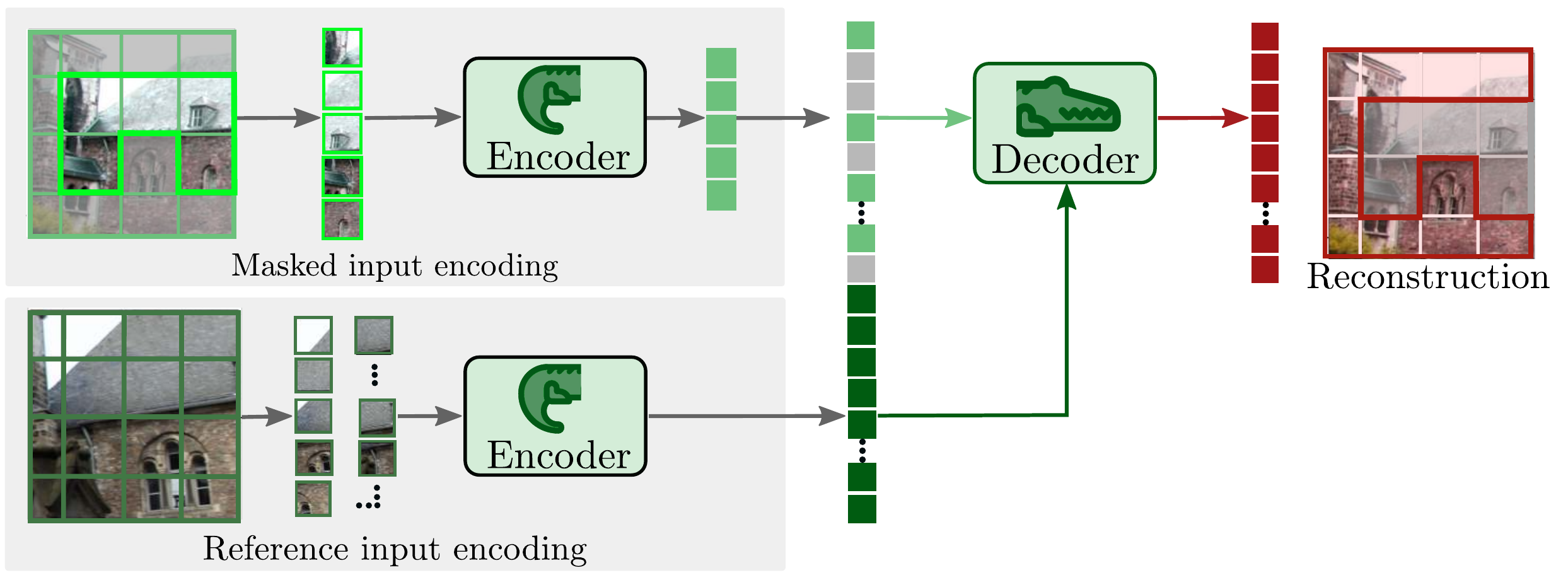} \\
\caption{\textbf{Overview of our CroCo model during pre-training.} 
Patches from the first input image are partially masked; visible ones are encoded into their latent representations, then padded with masked tokens to account for hidden patches. The same encoder is used to encode the patches of the second image. The decoder receives the encoded tokens from both images and use them to reconstruct the masked parts of the first  
image.}
\rsp{}
\label{fig:croco}
\end{figure}

Figure~\ref{fig:croco} provides an overview of our self-supervised model during pre-training.
We divide both images into sets of non-overlapping patches, denoted as tokens. Most tokens from the 
first image are randomly discarded, and the remaining ones are fed to an image encoder, which we implement using a Vision Transformer (ViT) backbone~\cite{DosovitskiyICLR21ViT}. All patches from the second (reference) image are encoded in a \emph{Siamese} manner, \ie, using the same encoder with shared weights. 
The token latent representations output by the encoder from both images, including tokens to account for the masked patches of the first 
image, are then fed to a decoder whose goal is to predict the appearance of hidden patches. 
To this aim, we use a series of transformer decoder blocks comprising cross-attention layers.
This allows non-masked tokens from the first image to attend 
tokens from the reference image, thus enabling cross-view comparison and reasoning.
Our model is trained using a simple pixel reconstruction loss over all masked patches, similar to MAE~\cite{he2021mae}.
To finetune the model on downstream tasks that process only a single image, \eg monocular depth estimation, the decoder can be discarded and we use the pre-trained encoder alone. In the case of binocular tasks such as optical flow estimation, 
the original CroCo architecture is used as is.

For pre-training, CroCo relies on pairs of images depicting the same scene; in this work the pairs were obtained from synthetic renderings of indoor scenes produced with the Habitat~\cite{habitat19iccv} simulator.
We empirically show that high masking ratios, \eg 90\%, lead to the best pre-training performance.
Our model is evaluated on monocular downstream tasks, in particular depth estimation on the NYUv2 dataset~\cite{nyuv2} and a diverse set of dense 2D and 3D regression tasks taken from Taskonomy~\cite{zamir2018taskonomy}.
We show that CroCo leads to significantly better performance compared to existing MIM models, whether they were pre-trained on the same data or on ImageNet~\cite{ILSVRC15imagenet}. 
When evaluated on monocular high-level semantic tasks, such as ImageNet classification, CroCo obtains lower performance compared to established MIM models, however.
This essentially comes from our use of indoor scenes 
for pre-training, instead of highly semantic datasets like ImageNet, as empirically shown in our experiments.
Lastly, we demonstrate that CroCo can be applied to binocular downstream tasks in a straightforward manner without bells and whistles. For instance, optical flow estimation can be performed by directly regressing 2 values per pixel using the decoder, and likewise, relative pose regression is achieved by simply appending a pose regression head to CroCo. 
In both cases, we show that CroCo pre-training leads to competitive results in these 3D vision downstream tasks.

\section{Related work}
\label{sec:related}

\PARp{Self-supervised representation learning}
is a paradigm developed to learn visual features from large-scale sets of unlabeled data \citep{JingPAMI21SelfSupervisedVisualFeatureLearningDeepSurvey}, before finetuning the model on downstream tasks, \eg classification or detection.
To achieve that, a \emph{pretext} task is designed and exploits inherent data attributes to automatically generate surrogate labels.
The earliest principle behind most existing pretext tasks for computer vision revolves around purposefully removing \emph{some} information from an image, \eg the color, the orientation or the ordering of a sequence of patches obtained from the image that the model then learns to recover~\citep{DosovitskiyNIPS14DiscriminativeUnsupervisedFeatureLearningCNN,GidarisICLR18UnsupervisedReprLearningPredictingImageRotations,NorooziECCV16UnsupervisedLearningVisReprSolvingJigsawPuzzles,NorooziCVPR18BoostingSelfSupervisedLearningViaKnowledgeTransfer}. 
It has been shown that despite the lack of any semantic supervision, networks trained to recover this artificially removed information learn useful representations and facilitate the training of various supervised downstream tasks~\citep{AsanoICLR20ACriticalAnalysisSelfSupervisionWhatLearnFromSingleImage,EricssonCVPR21HowWellDoSelfSupervisedModelsTransfer}.

Recently, the paradigm of instance discrimination has received a lot of attention, achieving highly competitive results in self-supervised visual representation learning~\cite{assran2022masked,CaronNIPS20SwAV,CaronICCV21DINO,ChenICML20SimCLR,grill2020byol,HeCVPR20MoCo}. It seeks to learn outputs that are invariant to well designed classes of data augmentation. Positive image pairs, obtained from the same instance by data augmentation, are pulled 
closer by the model, while samples obtained from different instances are pushed apart in the embedding space.

More recently, motivated by the success of BERT~\cite{DevlinNAACL19BERT} in NLP and by the introduction of Vision Transformers (ViT)~\cite{DosovitskiyICLR21ViT}, a variety of masked image prediction methods for self-supervised pre-training of vision models has been proposed. 
Reminiscent of denoising autoencoders~\cite{VincentJMLR10StackedDenoisingAutoencoders} or context encoders~\cite{PathakCVPR16ContextEncoders}, and aiming to reconstruct masked pixels \cite{AtitoX21SiTSelfSupervisedVIsionTransformer,ChenICML20GenerativepretrainingFromPixels,DosovitskiyICLR21ViT,ElNoubyX21AreLargeScaleDatasetsNecessarySelfSupervisedpretraining,he2021mae,XieX21SimMIM}, discrete tokens~\cite{BaoICLR22BEiT,ZhouICLR22IBOT} or deep features~\cite{BaevskiX22data2vecAGeneralFrameworkSelfSupervisedLearningSpeechVisionLanguage,WeiX21MaskedFeaturePrediction4SelfSupervisedVisualpretraining}, these methods have demonstrated the ability to scale to large datasets and models and achieve state-of-the-art results on various downstream tasks. In particular, the masked autoencoder (MAE)~\cite{he2021mae} accelerates pre-training by using an asymmetric architecture that consists of a large encoder that operates only on unmasked patches followed by a lightweight decoder that reconstructs the masked patches from the latent representation and mask tokens.
MultiMAE~\cite{bachmann2022multimae} leverages the efficiency of the MAE approach and extends it to multi-modal and multitask settings. Rather than masking input tokens randomly, the Masked Self-Supervised Transformer model (MST)~\cite{LiNIPS21MST} proposes to rely on the attention maps produced by a teacher network to dynamically mask low response regions of the input, and  a student network is then trained to reconstruct it.

\PAR{Self-supervised pre-training for dense downstream tasks}
Seminal self-supervised methods~\cite{chen2020simclr2,grill2020byol} were designed to output global image representations,
thus encoding limited local information, and
thereby hindering transferability of the learned models to downstream tasks involving dense per-pixel predictions, such as semantic segmentation. 
To alleviate this issue, several self-supervised methods have been proposed to perform contrastive learning on dense local representations rather than global ones~\cite{LiuX20SelfEMD,WangCVPR21DenseCL,XieCVPR21PixPro}.
In contrast, InsLoc~\cite{YangCVPR21InsLoc} proposes to paste image instances at various locations and scales onto background images, then predicts instance categories and foreground bounding boxes in the composed images. 
In VADeR~\cite{PinheiroNIPS20VADeR} and FlowE~\cite{XiongICCV21FlowE}, dense image representations are learned for several dense semantic downstream tasks such as object detection and semantic segmentation, using multi-hierarchy features~\cite{Wang2022RePre},
while~\cite{Yuan2021HRformer} combines multi-resolution parallel modules
with local-window self-attention to improve the memory and computation efficiency of dense prediction.
These self-supervision methods 
lead to improved performance on dense semantic downstream tasks such as object detection or semantic/instance segmentation. 
However, their design and evaluation protocols are not focused on 3D vision and lower-level geometric tasks
 such as depth, motion and flow estimation.
 
 For such tasks, the self-supervision signal can be obtained via view synthesis, 
 where the model is trained by enforcing photometric consistency~\cite{GodardCVPR17UnsupervisedMonocularDepthEstimationLeftRightConsistency,MeisterAAAI18UnFlowUnsupLearOpticalFlowBidirectionalCensusLoss,YangAAAI18UnsuperLearningGeometryEdgeAwareDepthNormalConsistency,ZhouCVPR17UnsupervisedLearningDepthEgoMotionFromVideo}, or from  RGB-D data,
where the model is trained to match 2D visual features with 3D geometric representations~\cite{BYOC,Pri3D,P4Contrast}.

\section{Cross-view Completion Pre-training}
\label{sec:pretrain}

In this section we present CroCo, our self-supervised pre-training  method based on cross-view completion and tailored to 3D vision tasks.

\PAR{Overview}
Figure~\ref{fig:croco} gives an overview of the CroCo architecture. 
Let $\bm{x_1}$ and $\bm{x_2}$ be two images of the same scene taken from different viewpoints.
Both images are divided into $N$ non-overlapping patches $\bm{p_1}= \{\bm{p_1}^1,\hdots,\bm{p_1}^{N}\}$, $\bm{p_2} = \{\bm{p_2}^1,\hdots, \bm{p_2}^{N}\}$, and a number $n=\lfloor rN \rfloor$ of tokens from the first set $\bm{p_1}$ is randomly masked, with $r\in[0,1]$ being a masking ratio hyper-parameter. 
We typically use $r=0.9$, \ie, 90\% of patches from $\bm{p}_1$ are discarded.
We denote the remaining set of visible patches from the first image by 
$\bm{\tilde{p}_1} = \{\bm{p_1}^i | m_i=0\}$, with $m_i=0$ indicating that the patch $\bm{p_1}^i$ is not masked ($m_i=1$ otherwise). 

An encoder $\mathcal{E_\theta}$ processes $\bm{\tilde{p}_1}$ and $\bm{p_2}$ independently, and a decoder $\mathcal{D}_\phi$ takes the encoding $\mathcal{E_\theta}(\bm{\tilde{p}_1})$, conditioned on encoding $\mathcal{E_\theta}(\bm{p}_2)$, in order to reconstruct $\bm{p}_1$:
\begin{equation}
\bm{\hat{p}_1} = \mathcal{D}_{\phi}\left(\mathcal{E_\theta}(\bm{\tilde{p}_1});\mathcal{E_\theta}(\bm{p}_2)\right).
\end{equation}

\PAR{Details on the encoder \bf{$\mathcal{E}_\theta$}}
A Siamese network denoted by $\mathcal{E}_\theta$ with shared weights $\theta$, implemented as a ViT~\cite{DosovitskiyICLR21ViT},
serves to encode the two input images previously split into independent patch sets $\bm{\tilde{p}_1}$ and $\bm{p_2}$.
In this study, we consider images of resolution $224 \times 224$, with a patch size of $16 \times 16$ pixels.
Following ViT, the encoder consists in a linear projection of the flattened input RGB patches, to which sinusoidal positional embeddings are added, followed by a series of transformer blocks~\cite{Waswani2017attention}, each composed of a multi-head self-attention block and an MLP (Multi-Layer Perceptron). 

\PAR{Details on the decoder \bf{$\mathcal{D}_\phi$}}
The decoder $\mathcal{D}_\phi$ with weights $\phi$ takes as input the set of encoded tokens from the first image $\mathcal{E}_\theta(\bm{\tilde{p}_1})$ concatenated with a learned representation $\bm{e}_{\text{mask}}$ that is repeated $n$ times to account for the masked patches. These tokens are iteratively processed by an attention-based block that also takes into account the encoded tokens $\mathcal{E}_\theta(\bm{p_2})$ from the reference image $\bf{x}_2$. A sinusoidal positional encoding is added to all tokens before being decoded by $\mathcal{D}_\phi$.
We experiment with two different architectures for the decoder block that are illustrated in Figure~\ref{fig:decoder}, see Appendix~\ref{sup:attn} for detailed equations.
The first one, termed~\textit{CrossBlock}, consists of (a) multi-head self-attention on the tokens representing the first image; (b) multi-head cross-attention between these tokens and the tokens in $\mathcal{E}_\theta(\bm{p_2})$ from the reference image, and (c) an MLP. 
The second architecture, termed \textit{CatBlock}, proceeds by concatenating the tokens from the two images, with an added learnable embedding for each of the two images, and then feeding these tokens to a series of standard transformer blocks, \ie, each consisting of (a) multi-head self-attention and (b) an MLP layer. Only the subset of tokens corresponding to the first image is taken into account for the final prediction.
These two attention blocks represent a trade-off between model size and computational cost. Indeed,~\textit{CrossBlock} has more learnable parameters, due to the cross-attention module, but a lower computational cost as it avoids the quadratic complexity of computing self-attention over the joint token set of size $2N$. 

\begin{figure}
\centering
\includegraphics[width=1\linewidth]{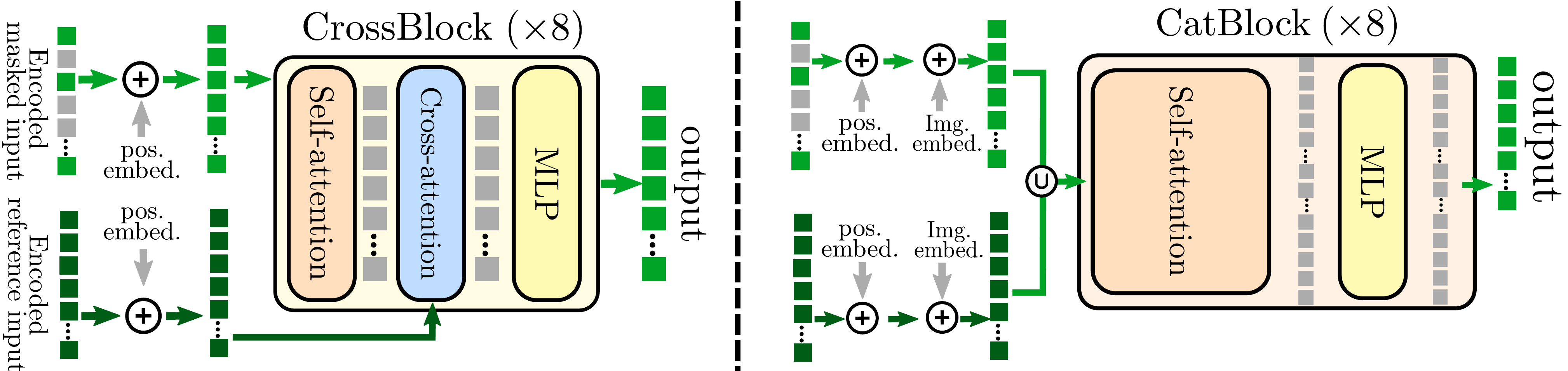} \\
\caption{\textbf{Attention blocks in the decoder.} CrossBlock (left) combines information from the input sets by alternating self-attention and cross-attention, while CatBlock (right) concatenates the two sets before applying self-attention.}
\rsp{}
\label{fig:decoder}
\end{figure}

\PAR{Training the CroCo model}
The decoder produces one feature vector per patch token, which is processed by a fully-connected layer that outputs 3 values (R, G and B) per pixel for each patch, \ie, $16 \times 16 \times 3 = 768$ values for squared patches with side $16$. These outputs serve as predicted reconstruction $\bm{\hat{p}_1}$, evaluated using a reconstruction error such as the Mean Squared Error (MSE) loss between predictions and ground-truth values of the pixels, averaged over the set of all masked tokens, denoted $\bm{p_1} \backslash \bm{\tilde{p}_1}$. 
Thus to perform self-supervised pre-training, the network is trained to minimize:
\begin{equation}
\mathcal{L}(\bm{x_1}, \bm{x_2}) = \frac{1}{\vert\bm{p_1} \backslash \bm{\tilde{p}_1}\vert} \sum_{\bm{p_1^i} \in \bm{p_1} \backslash \bm{\tilde{p}_1}} \Vert \bm{\hat{p}_1^i} - \bm{p_1^i}\Vert^2.
\end{equation}
We additionally try a variant of the loss where
each target patch is normalized using the mean and standard deviation of all pixel values within this patch.

\PAR{Pre-training details}
We implement our CroCo model in PyTorch~\cite{PaszkeNIPS19PyTorch} and train the network for 400 epochs using the AdamW optimizer~\cite{LoshchilovICLR19AdamW}.
We use a cosine learning rate schedule with a base learning rate of $1.5 \times 10^{-4}$ for an effective batch size of 256; with a linear warmup in the first 40 epochs. 
As encoder, we use a ViT-Base/16 backbone, \ie, a series of 12 transformer blocks with 768 dimensions and 12 heads for self-attention with patches of size $16 \times 16$. 
For both \textit{CrossBlock} and \textit{CatBlock} decoders, we use a series of 8 blocks with 512 dimensions and 16 attention heads. 
To account for the different dimensions of the encoder and the decoder, we apply a fully-connected layer between them. 

\textbf{Pre-training data.} We train our model on a set of 
synthetic image pairs of 3D indoor scenes derived from the HM3D~\cite{ramakrishnan2021hm3d}, ScanNet~\cite{dai2017scannet}, Replica~\cite{replica19arxiv} and ReplicaCAD~\cite{szot2021habitat} datasets as follows. 
In each 3D scene, we randomly sample up to 1000 pairs of camera viewpoints with a co-visibility greater than 50\%, and render these pairs of images using the Habitat simulator~\cite{habitat19iccv}.
In total, we generated a dataset of 1,821,391 pairs, that we refer to as~\emph{Habitat} in this paper.

\textbf{Finetuning CroCo.}
Our CroCo model can be used as pre-training to solve either monocular or binocular tasks, dense or not. 
For monocular task, the ViT-Base encoder is used alone, while binocular tasks benefit from the pre-training of both the encoder and the decoder.

\section{Experimental results} 
\label{sec:xp}

We evaluate CroCo on both monocular downstream tasks in Section~\ref{sub:monocular}, for which we provide ablations (Section~\ref{sub:ablate}) and a comparison to the state of the art (Section~\ref{sub:sota}), and on binocular tasks in Section~\ref{sub:binocular}. Training details for each task are available in the appendix.

\subsection{Monocular transfer tasks}
\label{sub:monocular}

\PAR{High-level semantic tasks}
While targeting 3D vision tasks, we still evaluate some high-level semantic tasks, namely \textit{image classification with linear probing} on ImageNet-1K~\cite{ILSVRC15imagenet} and report top-1 accuracy. We follow the exact same protocol as MAE~\cite{he2021mae} for that, with global average pooling for CroCo as we did not include a [CLS] token in our model.
Additionally, we report results on \textit{semantic segmentation} on ADE20k~\cite{ade} with 150 classes and 20,210 training images. We follow the protocol of~\cite{bachmann2022multimae} and report mean Intersection-over-Union (mIoU) on the validation set, using the same ConvNext~\cite{convnext} prediction head on top of the encoder.

\PAR{3D vision tasks} 
We evaluate \textit{monocular depth prediction} on  NYUv2~\cite{nyuv2} (795 training and 655 test images) by reporting $\delta_1$, \ie, the percentage of pixels that have an error ratio (max over prediction divided by ground-truth and its inverse) below $1.25$. We additionally report results on the Taskonomy dataset~\cite{zamir2018taskonomy} (800 training images) and report L1-loss on the `tiny' test set (54,514 images) for the same 8 tasks as~\cite{bachmann2022multimae}: \textit{curvature, depth, edges, 2D keypoints, 3D keypoints, normal, occlusion and reshading}. For better clarity with decimal number, we report the L1-losses multiplied by 1000.
 We also report the average ranking (rank.) over the 8 tasks in tables.
 For these tasks, we use a DPT head~\cite{DPT} on top of the encoder. We additionally experiment on the monocular task of absolute camera pose regression in Appendix~\ref{supsub:apr} and the binocular task of stereo image matching in Appendix~\ref{ssec:stereo}.
 
\subsubsection{Ablations}
\label{sub:ablate}

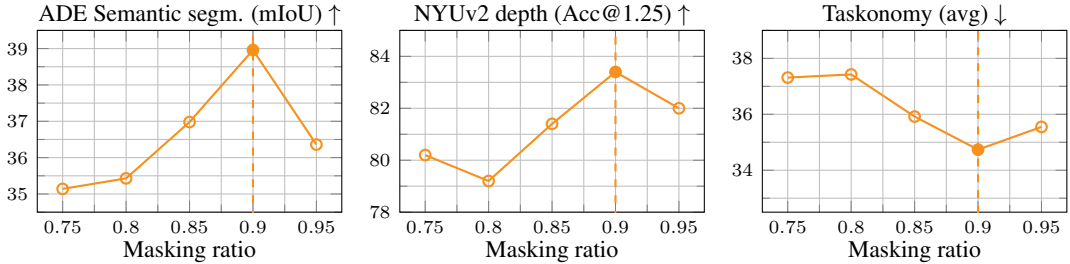
\begin{figure}[t]
\centering
\begin{subfigure}{.31\linewidth}
\hspace*{-0.3cm}
\input{plots/seg_f_mask}
\end{subfigure}
\hfill
\hspace{-0.2cm}
\begin{subfigure}{.31\linewidth}
\hspace*{-0.3cm}
\input{plots/depth_f_mask}
\end{subfigure}
\hfill 
\hspace{-0.2cm}
\begin{subfigure}{.31\linewidth}
\hspace*{-0.3cm}
\input{plots/taskonomy_f_mask}
\end{subfigure}
\hfill
\\[-0.3cm]
\caption{\textbf{Impact of the masking ratio} using the CrossBlock decoder and a loss on RGB values.}
\label{fig:masking_ratio}
\end{figure}

\PAR{Masking ratio}
We first report in Figure~\ref{fig:masking_ratio} the performance on semantic segmentation (ADE), depth estimation (NYUv2) and Taskonomy when varying the masking ratio $r$ using a \textit{CrossBlock} decoder and a loss without normalization. 
We observe that the best performance is obtained with a masking ratio of $r=90\%$.
This optimal ratio is higher than what was found for instance in MAE~\cite{he2021mae} for the auto-completion task. 
We attribute this to the help provided by the reference image in cross-view completion.
We show some reconstruction examples in Figure~\ref{fig:reconstructions}. 
Despite the small number of visible patches, this amount, combined with the reference view, seems to provide sufficient information to reconstruct the first image.
Note that the reconstructions tend to be blurry. This is due to the MSE loss, but as noted in MAE~\cite{he2021mae}, beyond a certain point, sharper reconstructions do not necessarily lead to better pre-training.

\begin{table}[t]
\centering
\begin{tabular}{cccccccc}
\toprule
Normalized & Decoder & ADE $\uparrow$ & NYUv2 $\uparrow$ & \multicolumn{2}{c}{Taskonomy $\downarrow$} & \multirow{2}{*}{FLOPs} & \multirow{2}{*}{Params} \\
\cmidrule(lr){3-3} \cmidrule(lr){4-4} \cmidrule(lr){5-6}
Target & Block& segm. & depth & avg. & rank. & & \\
\cmidrule(lr){1-2} \cmidrule(lr){3-6} \cmidrule(lr){7-8}
             & CrossBlock        & 39.0 & 83.4      & 34.74 & 2.50  & 50.2G & 120M \\
$\checkmark$ & CrossBlock        & 40.6       & 85.6      & \bf{33.00} & \bf{1.63}  & 50.2G & 120M \\
$\checkmark$ & CatBlock          & \bf{41.3}  & \bf{86.2} & 33.35   & 1.88   & 58.5G & 111M \\
\bottomrule
\end{tabular}
\caption{
\textbf{Impact of normalizing targets and decoder block} with a masking ratio of 90\%. 
We also indicate the number of FLOPs and parameters for the full network (encoder and decoder). 
\rsp{}
}
\label{tab:ablate}
\end{table}

\PAR{Normalized targets}
Regressing RGB values normalized by the mean and standard deviation within each patch has proven to be effective for MAE~\cite{he2021mae}. The first two rows of Table~\ref{tab:ablate} compare the results with or without this normalization for CroCo.
We observe that it indeed consistently improves the performance on all tasks, and we therefore use it in the remainder of the experiments.

\PAR{Decoder architecture}
In Table~\ref{tab:ablate}, we compare decoders built with the two proposed attention blocks: CrossBlock which uses cross-attention to exchange information between the two images, and CatBlock which concatenates the tokens from both images. Note that CatBlock has a smaller number of learnable parameters but a higher number of FLOPs. 
The performances obtained with the two proposed architectures are quite similar; CatBlock yielding slightly better performance for semantic segmentation or depth estimation while  CrossBlock performs better on 
Taskonomy. 
We favor the use of CrossBlock in what follows.
In Appendix~\ref{sup:decdepth}, 
we provide an ablation study on the number of blocks used in the decoder.

\PAR{Leveraging the decoder} 
Note that the pre-trained decoder can also be used when finetuning the model on monocular tasks: it suffices to feed the input image twice to the whole CroCo network, which in practice is done by duplicating the encoder output before passing it to the decoder. 
While this significantly increases the computational cost of the model, it yields a consistent gain of performance from $40.6$ to $41.0$ on ADE, {$86.1$} to $88.1$ on NYUv2,
and an improvement on $6$ out of $8$ tasks on Taskonomy, see further details in Appendix~\ref{supsub:decoder}. 

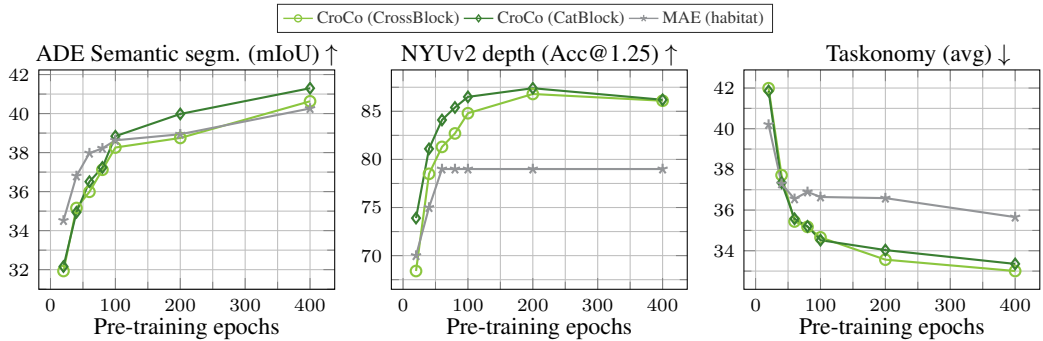
\begin{figure}
\centering
\begin{subfigure}{.8\linewidth}
\input{plots/legend_f_epoch}
\end{subfigure}  \\
\begin{subfigure}{.329\linewidth}
\input{plots/seg_f_epoch}
\end{subfigure}
\hfill
\begin{subfigure}{.329\linewidth}
\input{plots/depth_f_epoch}
\end{subfigure}
\hfill
\begin{subfigure}{.329\linewidth}
\input{plots/taskonomy_f_epoch}
\end{subfigure} \\
\caption{\textbf{Performance as a function of the number of pre-training epochs} for various models.}
\label{fig:profile}
\end{figure}

\PAR{Training profile}
In Figure~\ref{fig:profile} we show the performance on the downstream tasks as the pre-training progresses up to 400 epochs, when using CrossBlock or CatBlock in the decoder. 
We observe that the performance on 3D vision tasks -- depth estimation on NYUv2 and Taskonomy -- reaches a plateau and thus we evaluate our model pre-trained for $400$ epochs in what follows. 
We additionally train a MAE model\footnote{\url{https://github.com/facebookresearch/mae}} on the Habitat dataset, using all available images (\ie, twice the number of pairs). 
When pre-training on Habitat, both models achieve similar performance on the task of semantic segmentation, while CroCo significantly outperforms MAE on
depth prediction and Taskonomy.
This shows the benefit of using cross-view completion pre-training, rather than pure auto-completion, for 3D vision downstream tasks.

\PAR{Training pairs} CroCo pre-training requires pairs of images, which we obtain in our experiments by sampling two different points of view using the Habitat simulator. As an alternative, we evaluate using geometric transformations (homographies, rotations, scaling, crop, \etc) applied to an input image in order to generate the reference view; we report the results in Table~\ref{tab:selfpair} after 400 epochs of pre-training on Habitat. We observe a clear drop in performance on all downstream tasks, \ie, semantic segmentation, monocular depth estimation and Taskonomy. One  hypothesis is that in this case -- when there exists a synthetic transform between the two images -- the model can solve the pretext task of cross-view completion 
by directly fitting this transform, without reasoning about the geometry of the scene. 

This in turn raises the question of how to sample optimal viewpoints when generating synthetic image pairs using the Habitat simulator.
To answer it, we study the relation between downstream performance and co-visibility in pre-training image pairs (\ie, how much visual content is shared between the two images) in Appendix~\ref{sup:overlap}.
In short, it shows that using image pairs with a co-visibility ratio of approximately 0.5 leads to the best pre-trained models, and that on the one hand, very little co-visibility between the two images is sub-optimal as the task boils down to auto-completion, and on the other hand , if the two images composing a pair overlap too much, performance drops as the pre-training task becomes trivial.

\begin{table}[t]
\centering
\begin{tabular}{lcccc}
\toprule
\multirow{2}{*}{image pairs from} & ADE $\uparrow$ & NYUv2 $\uparrow$ & \multicolumn{2}{c}{Taskonomy $\downarrow$} \\
\cmidrule(lr){2-2} \cmidrule(lr){3-3} \cmidrule(lr){4-5}
& segm. & depth & avg. & rank. \\
\cmidrule(lr){1-1} \cmidrule(lr){2-5}
two viewpoints & \bf{38.8} & \bf{86.8} & \bf{33.56} & \bf{1.25} \\
geometric transformations of one image & 27.0 & 66.1 & 47.81 & 1.75 \\
\bottomrule
\end{tabular}
\caption{\textbf{Comparison with artificial pairs.} We compare pairs obtained by geometric transformations of one image to pairs sampled from two different viewpoints.
\rsp{}
}
\label{tab:selfpair}
\end{table}

\subsubsection{Comparison to the state of the art}
\label{sub:sota}

\begin{table}
\centering
\resizebox{\linewidth}{!}{
\begin{tabular}{lccccccccccccc}
\toprule
\multirow{2}{*}{pre-training method (data)} & IN1K $\uparrow$ & ADE $\uparrow$ & NYUv2 $\uparrow$ & \multicolumn{10}{c}{Taskonomy $\downarrow$} \\
\cmidrule(lr){2-2} \cmidrule(lr){3-3} \cmidrule(lr){4-4} \cmidrule(lr){5-14} 
 & lin. & segm. & depth & curv. & depth & edges & kpts2d & kpts3d & normal & occl. & reshad. & avg. & rank. \\
\midrule
DINO~\cite{CaronICCV21DINO} (IN1K)          & \bf{78.2} & 44.7      & 66.8 & 43.04 & 38.42 & 3.80 & 0.16 & 45.85 & 65.71 & 0.57 & 115.02 & 39.07 & 5.00 \\
MAE~\cite{he2021mae} (IN1K)                 & \underline{68.0}      & \underline{46.1}      & 79.6 & 41.59 & 35.83 & \bf{1.19} & \underline{0.08} & 44.18 & \underline{59.20} & \bf{0.55} & 106.08 & 36.09 & \underline{2.13} \\
MutliMAE~\cite{bachmann2022multimae} (IN1K) & 60.2      & \bf{46.4} & \underline{83.0} & \underline{41.42} & 35.38 & 2.17 & \bf{0.07} & \underline{44.03} & 60.35 & 0.56 & 105.25 & 36.17  & 2.75 \\
MAE (Habitat)                                   & 32.5      & 40.3      & 79.0 & 42.06 & \underline{33.63} & 1.79 & \underline{0.08} & 44.81 & 59.76 & 0.56 & \underline{102.54} & \underline{35.65} & 2.88 \\
{\color{OliveGreen}\bf{CroCo}} (Habitat)                             & 37.0 & 40.6 & \bf{85.6} & \bf{40.91} & \bf{31.34} & \underline{1.74} & \underline{0.08} & \bf{41.69} & \bf{54.13} & \bf{0.55} & \bf{93.58} & \bf{33.00} & \bf{1.25} \\
\bottomrule
\end{tabular}
}
\caption{\textbf{Comparison to the state-of-the-art pre-training methods} on  semantic tasks (image classification on ImageNet-1K with linear probing and semantic segmentation on ADE) and 3D vision tasks (NYUv2, Taskonomy), with a ViT-Base/16 backbone. We indicate the pre-training data in parenthesis. Best and second best results are \textbf{bold} and \underline{underlined}.
\rsp{}
}

\label{tab:sota}
\end{table}

In Table~\ref{tab:sota}, we compare the proposed CroCo pre-training strategy to the state of the art on monocular tasks. In particular, we compare to DINO~\cite{CaronICCV21DINO}, a state-of-the-art self-supervised method based on instance discrimination, and to MIM methods with MAE~\cite{he2021mae} and MultiMAE~\cite{bachmann2022multimae}.
The latter uses extra data modalities, namely depth and semantic segmentation generated by off-the-shelf supervised models for pre-training. All three methods rely on the ImageNet-1K (IN1K) dataset~\cite{ILSVRC15imagenet} for pre-training. This leads to high performance on high-level semantic tasks compared to our approach.
Note that DINO, which outputs a global representation, performs best on ImageNet-1K, while MIM-based approaches, which already output dense patch-level predictions during pre-training, obtain better results on dense tasks such as semantic segmentation or depth.

To measure the impact of the pre-training dataset, we report the performance of a MAE model trained on Habitat, \ie, with exactly the same data and ViT-Base architecture as used for CroCo. In this case, the performance on semantic tasks largely drops compared to pre-training on ImageNet  
(\eg 68.0\% to 32.5\% in classification on IN1K, see Table~\ref{tab:sota}).
This confirms that the high performance on semantic tasks are mostly due to the use of IN1K for pre-training.
When evaluating 3D vision downstream tasks, such as depth estimation on NYUv2 or Taskonomy, CroCo significantly outperforms all other methods.
Interestingly, it even outperforms MultiMAE for depth estimation on NYUv2 with 85.6\% Acc@1.25 \vs 83.0\%, while other approaches are below 80\%. 
On Taskonomy, CroCo performs best on 6 tasks out of 8 and is second best on the two other tasks, with an average rank of 1.25/5. 

\subsection{Applications to binocular tasks}
\label{sub:binocular}

We now empirically demonstrate that our CroCo model achieves competitive performance in two binocular tasks, namely optical flow and relative pose estimation, without any task-specific design. 

\PAR{Optical Flow}
We treat optical flow as a straightforward regression task and do not change the pre-training architecture except for modifying the regression head to output two flow channels instead of 3 RGB color channels.
We finetune our network on the public 40,000 synthetic pairs from AutoFlow~\cite{autoflow}, using a simple MSE loss on $224 \times 224$ crops, for 100 epochs, without any data augmentation besides color jittering.
We evaluate on the MPI-Sintel~\cite{sintel} dataset ($1041$ image pairs in the train split), on the clean and final renderings, using the average endpoint error (AEPE) metric.
To test on high resolution $1024 \times 536$ images, we regularly sample overlapping tiles of size $224 \times 224$ at the same position in both input images. We regress the flow between each pair of corresponding tiles. To recombine flow values, at each pixel location in the final prediction we use the flow value predicted by the nearest tile (\ie, based on the tile center). 

In Table~\ref{tab:optflow} we compare performance for various pre-training strategies and datasets on the MPI-Sintel training set.
Without pre-training, the network is unable to learn anything useful, as shown by the large AEPE values, 
 despite thorough but unsuccessful hyper-parameter tuning. 
This may be due to the high difficulty of the task coupled with the limited amount of training data. 
The performance significantly improves when we initialize the encoder with an MAE pre-trained model. 
With a CroCo initialization of both encoder \emph{and} decoder weights, however,
we again observe a large improvement with
an average decrease of the error by almost 2 pixels \wrt the best MAE model on both clean and final renderings.
Interestingly, we find that even with just 2 blocks in the decoder, a model pre-trained with CroCo still outperforms an MAE model having 8 decoder blocks (3.59/4.35 AEPE on clean/final, resp., compared to 4.63/5.24 AEPE for MAE; see Appendix~\ref{supsub:flow} 
for a complete ablation).
These results clearly demonstrate the benefits from pre-training with the cross-view completion task. 

\begin{table}[t]
    \centering
    \resizebox{0.54\linewidth}{!}{
    \begin{tabular}{llcc}
    \toprule
    encoder                             & decoder         & \multicolumn{2}{c}{MPI-Sintel}                \\
    init.                               & init.           & clean                                 & final     \\
    \midrule
    random                              & random          & 18.81                                 & 18.97     \\
    MAE (IN1K)                      & random          & 4.68                                  & 5.16      \\
    MAE (Habitat)                       & random          & 4.63                                  & 5.24      \\
    {\color{OliveGreen}\bf{CroCo}} (Habitat) & {\color{OliveGreen}\bf{CroCo}} (Habitat) & \bf{3.00} & \bf{3.60} \\
    \bottomrule
\end{tabular}
}
\caption{\label{tab:optflow}\textbf{Optical flow results} on the training set of the MPI-Sintel dataset for various pre-training methods when finetuned on AutoFlow. 
\rsp{}
}
\end{table}

Our results (3.00/3.60 on clean/final, resp.) are slightly behind recent state-of-the-art approaches like PWC-Net\cite{pwcnet} (2.55/3.93), LiteFlowNet2~\cite{liteflownet2} (2.24/3.78) and RAFT~\cite{raft} (1.43/2.71), all trained on FlyingChairs~\cite{flyingchairs} and FlyingThings~\cite{flyingthings} combined. However, our CroCo model remains competitive for this task considering the simplicity of our approach. 
In particular, note that (a) our model is trained from limited data 
without any sophisticated data augmentation, (b) our architecture 
is generic and not specially designed for optical flow prediction, unlike~\cite{pwcnet,liteflownet2,raft} which all rely on 4D cost volumes, 
(c) regression at test time is simple and straightforward,
(d) the small $224\times 224$ resolution used at test time hinders our performance on large displacements, a problem which we leave to future work.
We refer to Appendix~\ref{supsub:flow} 
for additional ablative experiments and comparisons.

\PAR{Relative pose estimation}
We experiment on the relative pose regression (RPR) downstream task.
Given a pair of images, the goal is to predict the relative pose
of the camera with respect to a reference view. 
To do so, we feed to the CroCo model two images corresponding to normalized camera views, and we 
regress the rigid transformation $(\bm{R},\bm{t}) \in SO(3) 
\times \mathbb{R}^3$ between these two views using a differentiable Procrustes layer~\cite{bregier2021deepregression} on top of the prediction head to ensure that $\bm{R}$ is a rotation matrix. 
We finetune the RPR model with an MSE loss 
between the predicted relative pose $(\bm{R}, \bm{t})$ 
and a ground truth $(\hat{\bm{R}},\hat{\bm{t}})$:
$\| \bm{R} - \hat{\bm{R}}\|_F^2 + \lambda \| \bm{t} - \hat{\bm{t}}\|^2$, with $\lambda$ set to $100$ in practice. 
During training, we augment the training image pairs with random permutations, color jittering and virtual camera rotations.
We evaluate on the 7-scenes dataset~\cite{glocker2013_7scenes}, with a single model finetuned on the union of the official training sets of all scenes, and tested on the corresponding test splits. 
Images are rescaled and cropped to a $224 \times 224$ resolution 
for simplicity.
At test time, for each image of the test split we retrieve the nearest image from the training set according to their AP-GeM-18 descriptors~\cite{revaud2019apgem}, and we predict the pose relative to this retrieved image.

\newcommand{\td}{\textdegree{}}
\begin{table}[ttt]
\caption{\label{tab:relative_pose}\textbf{Relative pose estimation results} with the median camera position and orientation errors on 7-scenes. Finetuning a model pre-trained with CroCo achieves competitive results compared to existing methods directly regressing relative camera pose, without applying any fusion technique.
\rsp{}
}
\resizebox{\linewidth}{!}{
\footnotesize{}
\setlength{\tabcolsep}{2pt}
\begin{tabular}{llllllllr}
\toprule{}
Method / pre-training
                                        & chess              & fire               & heads              & office
								        & pumpkin            & redkitchen         & stairs             & Average \\
\midrule{}
RelocNet*~\cite{balntas2018relocnet}
                                        & 12cm, 4.14\td{}    & 26cm, 10.44\td{}   & 14cm, 10.5\td{}    & 18cm, 5.32\td{}
								        & 26cm, 4.17\td{}    & 23cm, 5.08\td{}    & 28cm, 7.53\td{}    & 21cm, 6.74\td{}\\
NC-EssNet*~\cite{zhou2020essnet}
                                        & 12cm, 5.63\td{}    & 26cm, 9.64\td{}    & 14cm, 10.66\td{}   & 20cm, 6.68\td{}
								        & 22cm, 5.72\td{}    & 22cm, 6.31\td{}    & 31cm, 7.88\td{}    & 21cm, 7.50\td{}\\
CamNet*$^\dagger{}$~\cite{DingICCV19CamNetRetrievalForReLocalization}
                                & \underline{4cm}, \textbf{1.73\td{}}     
                                & \textbf{3cm, 1.74\td{}}     
                                & \underline{5cm}, \textbf{1.98\td{}}     
                                & \underline{4cm}, \textbf{1.62\td{}}
						        & \textbf{4cm}, \textbf{1.64\td{}}  
						        & \textbf{4cm}, \textbf{1.63\td{}}  
						        & \textbf{4cm}, \textbf{1.51\td{}}  
						        & \textbf{4cm}, \textbf{1.69\td{}} \\
\midrule{}%
top1 AP-GeM-18
                                        & 27.9cm, 12.81\td{} & 40.4cm, 16.06\td{} & 21.6cm, 16.46\td{} & 37.5cm, 12.79\td{}
								        & 44.4cm, 12.58\td{} & 46.7cm, 13.92\td{} & 32.2cm, 14.59\td{} & 36cm, 14.2\td{} \\
\midrule{}
MAE (Habitat)
    & 13.2cm, 9.44\td{} 
    & 32.0cm, 15.10\td{} 
    & 16.0cm, 16.75\td{}
    & 24.8cm, 11.54\td{} 
    & 25.4cm, 10.62\td{} 
    & 29.4cm, 13.32\td{} 
    & 32.8cm, 14.88\td{}
    & 24.8cm, 13.09\td{} \\
{\color{OliveGreen} \bf{CroCo}} (Habitat)
    & \textbf{2.4cm}, \underline{2.81\td{}} 
    & \underline{4.0cm}, \underline{3.86\td{}} 
    & \textbf{3.1cm}, \underline{4.00\td{}} 
    & \textbf{3.4cm}, \underline{2.53\td{}}
    & \underline{4.9cm}, \underline{2.79\td{}} 
    & \underline{5.5cm}, \underline{3.72\td{}} 
    & \underline{11.7cm}, \underline{4.53\td{}} 
    & \underline{5.0cm}, \underline{3.46\td{}}  \\
\bottomrule{} \vspace{-0.2cm} \\
\multicolumn{9}{c}{*: fuse multiple pose predictions   ~~~~~~~~ $^\dagger{}$: exploit temporal information and multi-step retrieval}
\end{tabular}
}
\end{table}

For each scene, we report the standard median position and median orientation error in Table~\ref{tab:relative_pose}.
We compare results obtained with the model initialized with CroCo to that of a model with an encoder initialized with MAE (Habitat) and a decoder trained from scratch (last two rows). Unsurprisingly, the former leads to significantly better results than the latter.
We furthermore compare our model pre-trained with CroCo  to other state-of-the-art RPR methods which directly regress relative camera
poses~\cite{balntas2018relocnet, DingICCV19CamNetRetrievalForReLocalization} or essential matrices~\cite{zhou2020essnet} as well as the the retrieval baseline consisting in predicting the identity as relative pose (see Table~\ref{tab:relative_pose} upper part). Our model outperforms all methods except CamNet~\cite{DingICCV19CamNetRetrievalForReLocalization}. Note that the latter fuses multiple pose predictions, exploit temporal information and multi-step retrieval; in contrast our model directly predicts the relative pose from a (query, map) image pair, without any further processing, and is thus significantly simpler.

\section{Discussion}
\label{sec:discussion}

We have introduced the novel task of cross-view completion for pre-training computer vision networks tailored to 3D vision downstream tasks.
Our experiments show that cross-view completion allows to learn representations better suited to 
3D vision tasks than classical MIM with auto-completion, and straightforwardly transfer to monocular and binocular tasks.
A limitation of our model pre-trained with cross-view completion is that it seems less tailored to high-level semantic tasks.
This is arguably more related to the choice of the dataset than to the pre-training task itself, and future work could explore the use of cross-view completion in conjunction with more object-centric datasets like ImageNet.
Our approach requires pairs of images depicting the same scene, and in this work, we leverage synthetic renderings only. Future work could also extend it to real-world image pairs, which can be obtained without any supervision, for instance  
with Structure-from-Motion~\cite{megadepth,sfm120k} or geo-referencing. 

{\small
\bibliographystyle{plain}
\bibliography{macros,biblio}
}

\clearpage
\appendix

{\huge \bf{Appendix} \vspace{1cm}}

\def\Loss{\mathcal L}

This appendix is structured as follows. In Section~\ref{sup:acro}, we first 
provide  a list of acronyms used throughout the paper, followed by further
details about our decoder architectures (Section~\ref{sup:attn}).
In Section~\ref{sup:pre-train} we detail the parameter setting used in our pre-training and we provide an analysis about how the overlap between the training image pairs affects the pre-training and downstream performance (Section~\ref{sup:overlap}) as well as an ablation on the impact of the decoder depth (Section~\ref{sup:decdepth}).
Section~\ref{sup:monocular} then gives further details about monocular tasks, including a set of experiments on a complementary task, namely absolute pose regression  (Section~\ref{supsub:apr}). In Section~\ref{sup:binocular}, we give more details and visual examples concerning the binocular tasks as well as a complementary task, namely stereo matching  (Section~\ref{ssec:stereo}). 
Next, we list assets and computing resources in Section~\ref{sup:asset}.
Finally, we provide additional reconstruction examples (Section~\ref{sup:visualex}).

\section{List of acronyms}
\label{sup:acro}

We provide the list of acronyms used in the paper, excluding acronyms related to papers or datasets with direct references. 

\begin{center}
\begin{tabular}{ll}
\toprule 
acronym & meaning\tabularnewline
\midrule 
Acc@X & Accuracy at a certain error threshold X\tabularnewline
AEPE & Average EndPoint Error\tabularnewline
CroCo & Cross-view Completion\tabularnewline
DINO & DIstillation with NO labels \cite{CaronICCV21DINO}\tabularnewline
DPT & Dense Prediction Transformer \cite{DPT}\tabularnewline
FLOPs & Floating-Point Operations\tabularnewline
IN1K & ImageNet-1K \cite{ILSVRC15imagenet}\tabularnewline
mIoU & mean Intersection-over-Union\tabularnewline
MAE & Masked Auto-Encoder \cite{he2021mae}\tabularnewline
MIM & Masked Image Modeling\tabularnewline
MLP & Multi-Layer Perceptron\tabularnewline
MSE & Mean Squared Error\tabularnewline
NLP & Natural Language Processing\tabularnewline
RPR & Relative Pose Regression\tabularnewline
ViT & Vision Transformers \cite{DosovitskiyICLR21ViT}\tabularnewline
\bottomrule 
\end{tabular}
\end{center}

\section{Details on the decoder architectures}
\label{sup:attn}

In this section, we provide detailed equations about the blocks used in the CroCo model architecture.

For the encoder, we use a standard transformer block. Let $X \in \mathbb{R}^{N\times D}$ be the $N$ $D$-dimensional tokens as input to the transformer block. The transformer block computes:

\begin{eqnarray}
\bar{X} & = & \text{LayerNorm}(X) \nonumber \\ 
X'  & = & X+\text{Attention}\left(W_{Q}^{S}\bar{X},W_{K}^{S}\bar{X},W_{V}^{S}\bar{X}\right) \\
\text{Output} & = & X'+\text{MLP}(\text{LayerNorm}(X')) \nonumber,
\end{eqnarray}
where $W_{Q}^{S},W_{K}^{S},W_{V}^{S}$ are learnable parameters. In practice, a bias is also learned and applied to these 3 projections but are omitted in the equations for the sake of clarity. The attention itself is computed classically
as:

\begin{equation}
\text{Attention}(Q,K,V)= \text{Proj} \Big( \text{softmax}\left(\frac{QK^{\top}}{\sqrt{D}}\right)V \Big),
\end{equation}
where $\text{Proj}$ denotes a projection layer, \ie, a fully-connected layer.

We now provide equations for the two blocks we have tried in the decoder, namely the \emph{CrossBlock} and the \emph{CatBlock}.

In the \emph{CrossBlock} decoder architecture, self-attention and cross-attention are used one after the other. Let $X,Y\in\mathbb{R}^{N\times D}$ be the $N$ input  tokens  to the block from the two views respectively. The CrossBlock output is computed by:

\begin{eqnarray}
\bar{X} & = & \text{LayerNorm}(X) \nonumber \\ 
\bar{Y} & = & \text{LayerNorm}(Y) \nonumber \\
X'  & = & X+\text{Attention}\left(W_{Q}^{S}\bar{X},W_{K}^{S}\bar{X},W_{V}^{S}\bar{X}\right) \\
X'' & = & X'+\text{Attention}\left(W_{Q}^{C}\text{LayerNorm}(X'),W_{K}^{C}\bar{Y},W_{V}^{C}\bar{Y}\right) \nonumber\\
\text{Output} & = & X''+\text{MLP}(\text{LayerNorm}(X'')) \nonumber,
\end{eqnarray}

where $W_{Q}^{S},W_{K}^{S},W_{V}^{S}$ denote learnable parameters
for the self-attention and likewise $W_{Q}^{C},W_{K}^{C},W_{V}^{C}$
for the cross-attention.

For the \emph{CatBlock} decoder architecture, and following
the same notations, we first concatenate $X$ and $Y$ to form
$Z=\left[X+v_{1},Y+v_{2}\right]\in\mathbb{R}^{2N\times D}$ before
the first block, where $v_{1}$, $v_{2}\in\mathbb{R}^{D}$ are learnable
embeddings specifying the input view. The CatBlock output is computed by:

\begin{eqnarray}
\bar{Z}       & = & \text{LayerNorm}(Z) \nonumber \\
Z'            & = & Z+\text{Attention}\left(W_{Q}^{S}\bar{Z},W_{K}^{S}\bar{Z},W_{V}^{S}\bar{Z}\right) \\
\text{Output} & = & Z'+\text{MLP}(\text{LayerNorm}(Z')) \nonumber.
\end{eqnarray}

\section{Further details on cross-view completion pre-training}
\label{sup:pre-train}

\subsection{Detailed pre-training settings}

We report 
below the detailed parameter setting we used in our pre-training.

\begin{center}
\begin{tabular}{ll}
\toprule
Hyperparameters & Value \\
\midrule
Optimizer & AdamW~\cite{LoshchilovICLR19AdamW} \\
Base learning rate & 1.5e-4 \\
Weight decay & 0.05 \\
Adam $\beta$ & (0.9, 0.95) \\
Batch size & 256 \\
Learning rate scheduler &  Cosine decay \\
Training epochs & 400  \\
Warmup learning rate & 1e-6\\
Warmup epochs & 40 \\
\midrule
Masked tokens & 90\% \\
Input resolution & $224 \times 224$ \\
Augmentation & Homography, Color jitter \\
\bottomrule
\end{tabular}
\end{center}

\vspace{5mm}

\subsection{Ablation on overlaps between pre-training pairs}
\label{sup:overlap}

\begin{figure}[h]
    \centering
    \includegraphics[height=2.2in]{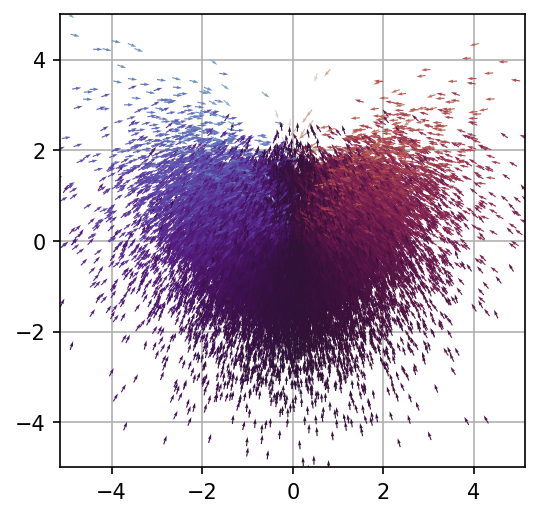}
    \includegraphics[height=2.3in]{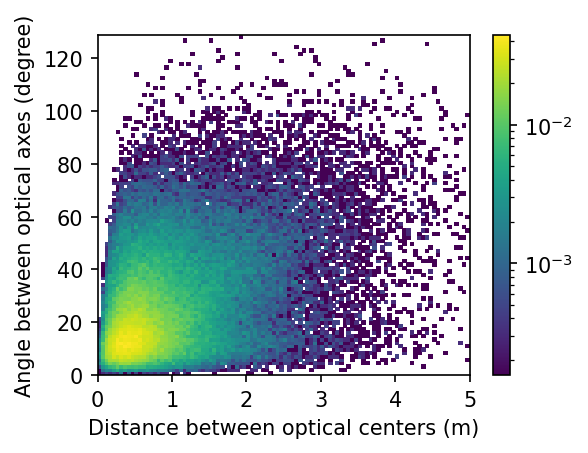}
    \caption{\textbf{Distribution of viewpoints pairs used for pre-training.} Left: 2D position (in meters) and orientation (arrow) of one view with respect to the other, once projected on the world horizontal plane.
    Right: Joint histogram of distances and angles within pairs of views used for training.}
    \label{fig:habitat2_statistics}
\end{figure}

For pre-training, we use synthetic pairs generated using the Habitat simulator, keeping pairs with a co-visibility ratio over $0.5$.
Figure~\ref{fig:habitat2_statistics} presents some statistics about the distribution of viewpoints considered.
Intuitively, the choice of this co-visibility threshold was guided by two important observations:
(a) if two images composing a pair overlap too little, the task boils down to auto-completion, therefore we set a threshold on the minimal co-visibility ratio of 0.5 in our main experiments, (b) if two images composing a pair overlap too much, the task becomes trivial, therefore we encourage large viewpoint changes between images in our training set.
In this section, we present an ablation to better understand what makes good pre-training pairs.

\begin{figure}[h]
    \centering
    \includegraphics[width=\linewidth]{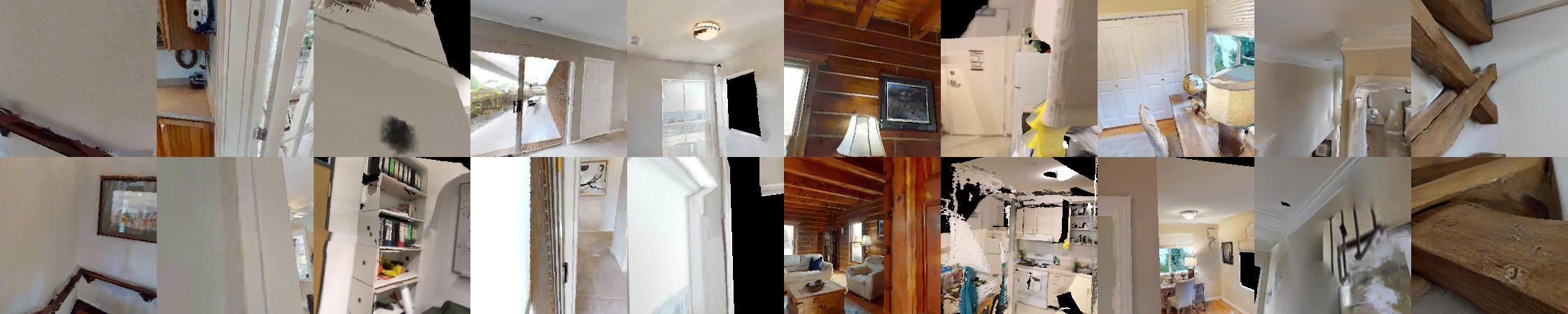}
    Co-visibility ratio between 0 and 0.1\\
    \vspace{5mm}
    \includegraphics[width=\linewidth]{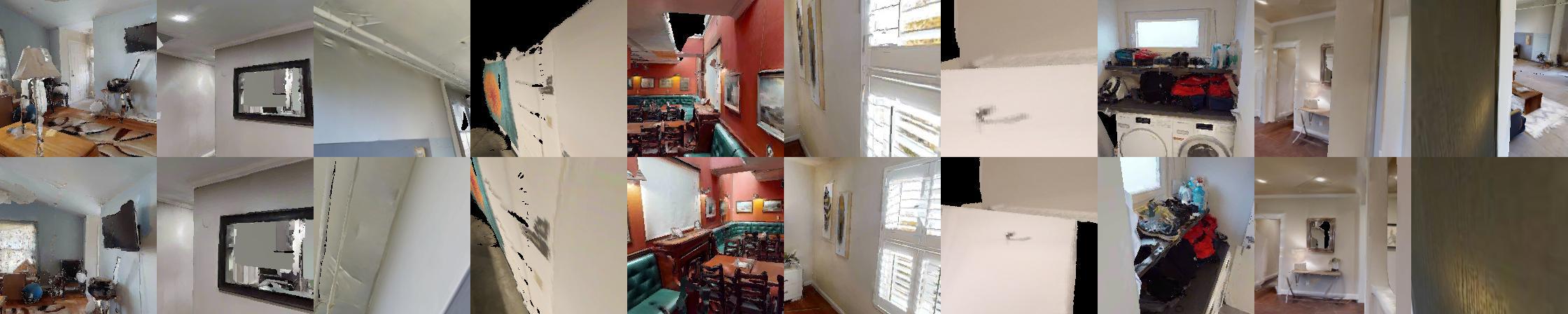}
    Co-visibility ratio between 0.4 and 0.5 \\
     \vspace{5mm}
    \includegraphics[width=\linewidth]{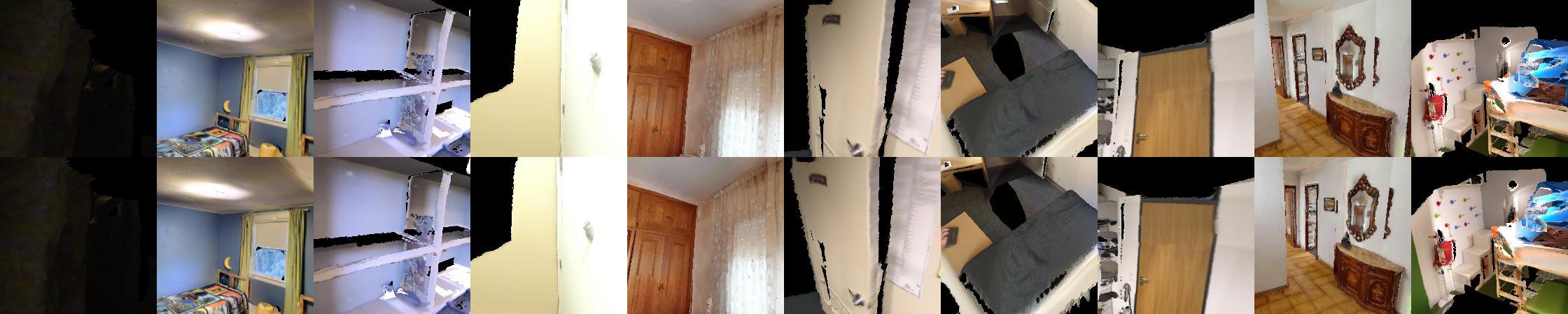}
    Co-visibility ratio between 0.9 and 1.0
    \caption{\textbf{Some pre-training pair examples for various co-visibility ratios.}}
    \label{fig:covisibility_bins_samples}
\end{figure}

\begin{figure}
    \centering
    \includegraphics[width=\linewidth]{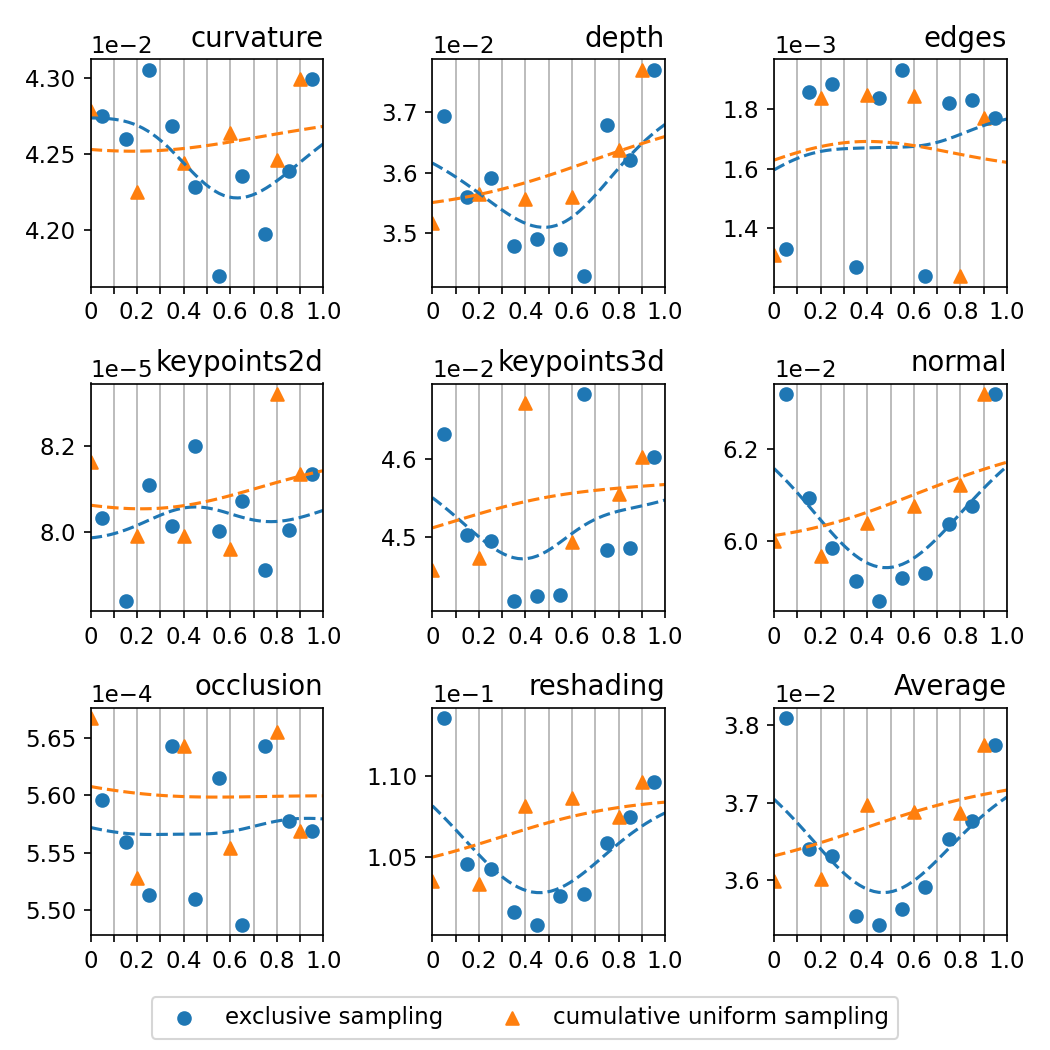}
    \caption{\textbf{Ablation on co-visibility between pre-training pairs}. Performance achieved on Taskonomy tasks (vertical axis, lower is better) by a CroCo model pre-trained with image pairs of co-visibility ratio exclusively distributed in the range $[x-0.05, x+0.05]$ (blue) or uniformly distributed in the range $[x, 1.0]$ (orange), for various values of $x$ (horizontal axis). Dashed lines represent trend curves obtained by Gaussian smoothing.}
    \label{fig:ablation_covisibility_taskonomy}
\end{figure}

Formally, we define the visibility ratio $v_{i,j}$ of a view $i$ with respect to an other view $j$ as the ratio of pixels of image $i$ that are visible in image $j$. We ignore pixels where no geometry is rendered by the Habitat simulator for this computation.
We similarly define the co-visibility ratio between two views $i$ and $j$ as $\min(v_{i,j}, v_{j,i})$.
To study the influence of co-visibility on pre-training, we generate multiple training sets each composed of $700,000$ image pairs with different co-visibility distributions, and pre-train CroCo on these sets for a number of steps equivalent to 200 epochs of the original dataset, while keeping all other parameters fixed.
Figure~\ref{fig:covisibility_bins_samples} provides examples of such training pairs. We report in Figure~\ref{fig:ablation_covisibility_taskonomy} the performance achieved when  pre-training with such pairs and then finetuning on Taskonomy tasks.
Overall, we observe better performance for most 3D-related tasks (in particular for \emph{curvature}, \emph{depth}, \emph{keypoints3d}, \emph{normal}, \emph{reshading}) when pre-training with pairs with a co-visibility ratio close to 0.5, compared to pre-training with pairs of greater or lower co-visibility (blue curve in Figure~\ref{fig:ablation_covisibility_taskonomy}).
We also experimented with pairs chosen to have co-visibility ratios uniformly distributed over different value ranges and report results in Figure~\ref{fig:ablation_covisibility_taskonomy}. We find that pre-training with co-visibility ratios uniformly distributed  over $[0, 1.0]$ leads to worse results than when exclusively sampling  pairs with a co-visibility ratio within $[0.4,0.5]$. This confirms that pairs having an intermediate co-visibility ratio are the most suitable for pre-training, although more extensive experiments would be required to strongly support these findings.

\clearpage

\subsection{Ablation on decoder depth}
\label{sup:decdepth}

In this section, we report an ablation on the decoder depth, \ie, varying the number of decoder blocks for a decoder using \textit{CrossBlock}. The results are shown in  Table~\ref{tab:decdepth}. 
Note that in the main paper we used a decoder with 8 blocks. 
We observe that the decoder depth has overall little impact on the performance for monocular tasks (semantic segmentation on ADE, depth prediction on NYUv2 or Taskonomy dense tasks). 
For the binocular task of optical flow estimation on MPI-Sintel, the error clearly decreases as the depth is increased, showing the importance of a deeper decoder when the decoder is also leveraged. Note however that even with a decoder of depth 2, which means that the network comparing the two images is extremely shallow, a competitive performance can still be reached for optical flow.
In particular, the performance with 2 decoder blocks is still largely superior to that of a MAE-pre-trained network having a deep decoder (8 blocks) finetuned in the same conditions. 

\begin{table}[h]
\centering
\begin{tabular}{cccccccrr}
\toprule
Decoder & ADE $\uparrow$ & NYUv2 $\uparrow$ & \multicolumn{2}{c}{Taskonomy $\downarrow$} & \multicolumn{2}{c}{MPI-Sintel $\downarrow$} & \multirow{2}{*}{FLOPs} & \multirow{2}{*}{Params} \\
\cmidrule(lr){2-2} \cmidrule(lr){3-3} \cmidrule(lr){4-5} \cmidrule(lr){6-7}
Depth & segm. & depth & avg. & rank. & clean & final & & \\
\cmidrule(lr){1-1} \cmidrule(lr){2-7} \cmidrule(lr){8-9}
2 & 38.9      & 84.6      & 33.83      & 2.50 & 3.59 & 4.35 & 39.3G & 94M \\
4 &  \underline{40.2}      & \bf{86.7} & \bf{32.80} & \bf{1.50} & 3.15 & 3.86 & 42.9G & 103M \\
6 & 38.7      & 83.6      & 34.14      & 3.25 &  \underline{3.09} &  \underline{3.79} & 46.6G & 111M \\
8 & \bf{40.6} &  \underline{85.6}      &  \underline{33.00}      &  \underline{1.88} & \bf{3l.00} & \bf{3.60} & 50.2G & 120M \\
\bottomrule
\end{tabular}
\caption{
\textbf{Impact of decoder depth} with CrossBlock. We used 8 blocks in the decoder in the main paper. Best result per column in bold and second best underlined.
}
\label{tab:decdepth}
\end{table}

\section{Further details and results for the monocular downstream tasks }
\label{sup:monocular}

\subsection{Detailed finetuning settings for monocular tasks}

We provide in this section details of the finetuning settings used for our monocular tasks experiments.
Linear probing settings are summarized below, 
Note that they correspond exactly to the settings presented by the authors of MAE~\cite{he2021mae} with the LARS~\cite{lars} optimizer and a large batch size.\\

\begin{center}
\begin{tabular}{ll}
\toprule
Hyperparameters & Value \\
\midrule
Optimizer & LARS~\cite{lars} \\
Base learning rate & 0.1 \\
Weight decay & 0 \\
Optimizer momentum & 0.9 \\
Batch size & 16,384 \\
Learning rate sched. &  Cosine decay \\
Warmup epochs & 10 \\
Training epochs & 90 \\
Augmentation & RandomResizeCrop \\
\bottomrule
\end{tabular}
\end{center}

\vspace{10mm}

We further detail the training settings for semantic segmentation on ADE20k, monocular depth estimation on NYUv2 and Taskonomy regression tasks.
These settings closely match those presented by the authors of MultiMAE~\cite{bachmann2022multimae}.\\

\begin{center}
\begin{tabular}{llll}
\toprule
Hyperparameters & ADE & NYUv2 & Taskonomy \\
\midrule
Optimizer & AdamW~\cite{LoshchilovICLR19AdamW} & AdamW~\cite{LoshchilovICLR19AdamW} & AdamW~\cite{LoshchilovICLR19AdamW} \\
Learning rate & 1e-4 & 3e-5& 2.5e-4 \\
Layer-wise lr decay 0.75 & - & 0.75 \\
Weight decay & 0.05 & 1e-6 & 2.5e-2 \\
Adam $\beta$ & (0.9, 0.999) & (0.9, 0.999) & (0.9, 0.999) \\
Batch size & 16 & 16 & 32 \\
Learning rate sched. & Cosine decay & Cosine decay & Cosine decay\\
Training epochs & 64 & 1500 & 300  \\
Warmup learning rate & 1e-6 & - & 1e-6 \\
Warmup epochs & 1 & 100 & 5 \\
\midrule
Input resolution & $512 \times 512$ & $256 \times 256$ & $384 \times 384$ \\
\multirow{2}{*}{Augmentation} & Large Scale jittering & RandomCrop &  \multirow{2}{*}{-} \\
 & Color jittering  & Color jittering & \\
Drop path & 0.1 & 0.0 & 0.1 \\
\bottomrule
\end{tabular}
\end{center}

\vspace{10mm}

\subsection{Leveraging the decoder for monocular tasks}
\label{supsub:decoder}

By default, for dense monocular tasks we append a DPT module~\cite{DPT} to the encoder of our CroCo model; this means that the decoder is discarded when finetuning. We now discuss other possible ways of finetuning our CroCo model on downstream tasks.
\begin{itemize}
\item \texttt{Using the encoder alone.} The simplest and most lightweight option is to use the encoder alone, without the decoder nor the DPT module, by appending an output linear prediction head, trained from scratch for the downstream task. In Table~\ref{tab:finetune} we see, by comparing rows 1 \& 2 and rows 5 \& 6, that removing the DPT module results in a significant degradation of performance across all tasks. 
\item \texttt{Decoder instead of the DPT module.} The decoder of our CroCo model can be used instead of the DPT module. While it does not fuse information from several layers at different depths, as the DPT module does, 
the decoder is trained to output dense predictions, and as such it should be easy to finetune for dense tasks other than RGB predictions. To achieve this we initialize the full CroCo model with pre-trained weights, duplicate the encoded features from the input image to serve as input to the decoder, and simply replace the prediction head of the network by a new one trained from scratch.
In Table~\ref{tab:finetune} we observe by comparing rows 2 \& 3 as well as 6 \& 7 that on most tasks, this leads to a very minor degradation of performance. Furthermore, we found that finetuning the model this way was one order of magnitude faster than training a DPT module: convergence on NYUv2 requires approximately $1500$ epochs for models without a decoder, against approximately $200$ for models using the decoder.
\item \texttt{Decoder and DPT module.} The decoder can be used together with the DPT module, by extracting the features to be used as input to the DPT module from the CroCo decoder rather than from the encoder. This increases the cost of running the model. However, in Table~\ref{tab:finetune} we see that this leads to consistent performance gains, by comparing rows 2 \& 4, or rows 6 \& 8. 
\item \texttt{Frozen backbones.} In all cases, we can choose to freeze the backbone and only train the new prediction layers (DPT module or output prediction head when no DPT module is used).
Such an experiment is useful to probe what information the pre-trained model captures.
It can also help with over-fitting in cases where very little data is used to train for the downstream task. Empirically, we observe by comparing the lower half of Table~\ref{tab:finetune} to the upper part that freezing the backbone degrades the performance in a majority of cases. On average, however, the performance remains surprisingly high, which demonstrates that frozen output features produced by the CroCo model are already meaningful representations for a wide diversity of 3D vision tasks.
\end{itemize}

\vspace{10mm}

\newcommand{\mbm}[1]{$\bm{#1}$}
\begin{table}[h]
    \centering
    \caption{ \textbf{Performance of our CroCo model when finetuned for monocular downstream tasks using different modules}, and possibly freezing the backbone network. Note that in the main paper, performance was reported without the decoder, with DPT and with finetuning of the backbone.}
    \label{tab:finetune}
    \resizebox{\linewidth}{!}{
  \begin{tabular}{ccccccccccccc}
	\toprule
	\multicolumn{3}{c}{architecture} & NYUv2 $\uparrow$ & \multicolumn{9}{c}{Taskonomy $\downarrow$} \\
	\cmidrule(lr){1-3} \cmidrule(lr){4-4} \cmidrule(lr){5-13} dec. & DPT & frozen & depth & curv. & depth & edges & kpts2d & kpts3d & normal & occl. & reshad. & avg. \\
	\midrule
	  \multicolumn{13}{c}{With finetuned backbone} \\
	  \midrule
	  \xmark & \xmark & \xmark & 67.8             & 45.67             & 63.56             & 10.50            & 0.51             & 60.20             & 139.69            & 0.65             & 186.6             & 63.42             \\ 
	  \xmark & \cmark & \xmark & \underline{86.1} & \underline{40.91} & \underline{31.34} & \underline{1.74} & \mbm{0.08}        & \underline{41.69} & \underline{54.13} & \mbm{0.55}        & \underline{93.58} & \underline{33.00} \\ 
	  \cmark & \xmark & \xmark & 85.9             & 45.05             & 33.93             & 4.18             & \underline{0.15} & 44.90             & 63.02             & \mbm{0.55}        & 103.1             & 36.86             \\ 
	  \cmark & \cmark & \xmark & \mbm{88.1}        & \mbm{39.93}        & \mbm{30.88}        & \mbm{1.66}        & 0.58             & \mbm{40.87}        & \mbm{52.26}        & \underline{0.56} & \mbm{90.77}        & \mbm{32.18}        \\ 

	  \midrule
	  \multicolumn{13}{c}{With frozen backbone}                                                   \\
	  \midrule
	  \xmark & \xmark & \cmark & 51.3             & 49.39             & 69.25             & 34.76            & 0.68             & 60.15             & 170.10     & 0.80             & 198.2             & 72.92                    \\ 
	  \xmark & \cmark & \cmark & 85.2             & \underline{42.01} & \mbm{38.18}        & \mbm{2.43}        & \mbm{0.09}        & \underline{45.50} & \mbm{64.58} & \underline{0.55} & \underline{117.4} & \underline{38.85}        \\ 
	  \cmark & \xmark & \cmark & \underline{86.4} & 44.73             & 75.97             & 35.16            & 0.56             & 60.54             & 177.12     & 0.65             & 220.0             & 76.84                    \\ 
	  \cmark & \cmark & \cmark & \mbm{87.1}        & \mbm{41.57}        & \underline{41.27} & \underline{3.64} & \underline{0.24} & \mbm{43.49}        & \mbm{62.77} & \mbm{0.54}        & \mbm{116.9}        & \mbm{38.81}               \\ 
	\bottomrule
\end{tabular}
}
\end{table}

\vspace{5mm}

\subsection{Visualization of monocular depth prediction  and Taskonomy results}

\begin{figure}[h] 
    \centering
    \includegraphics[width=\linewidth]{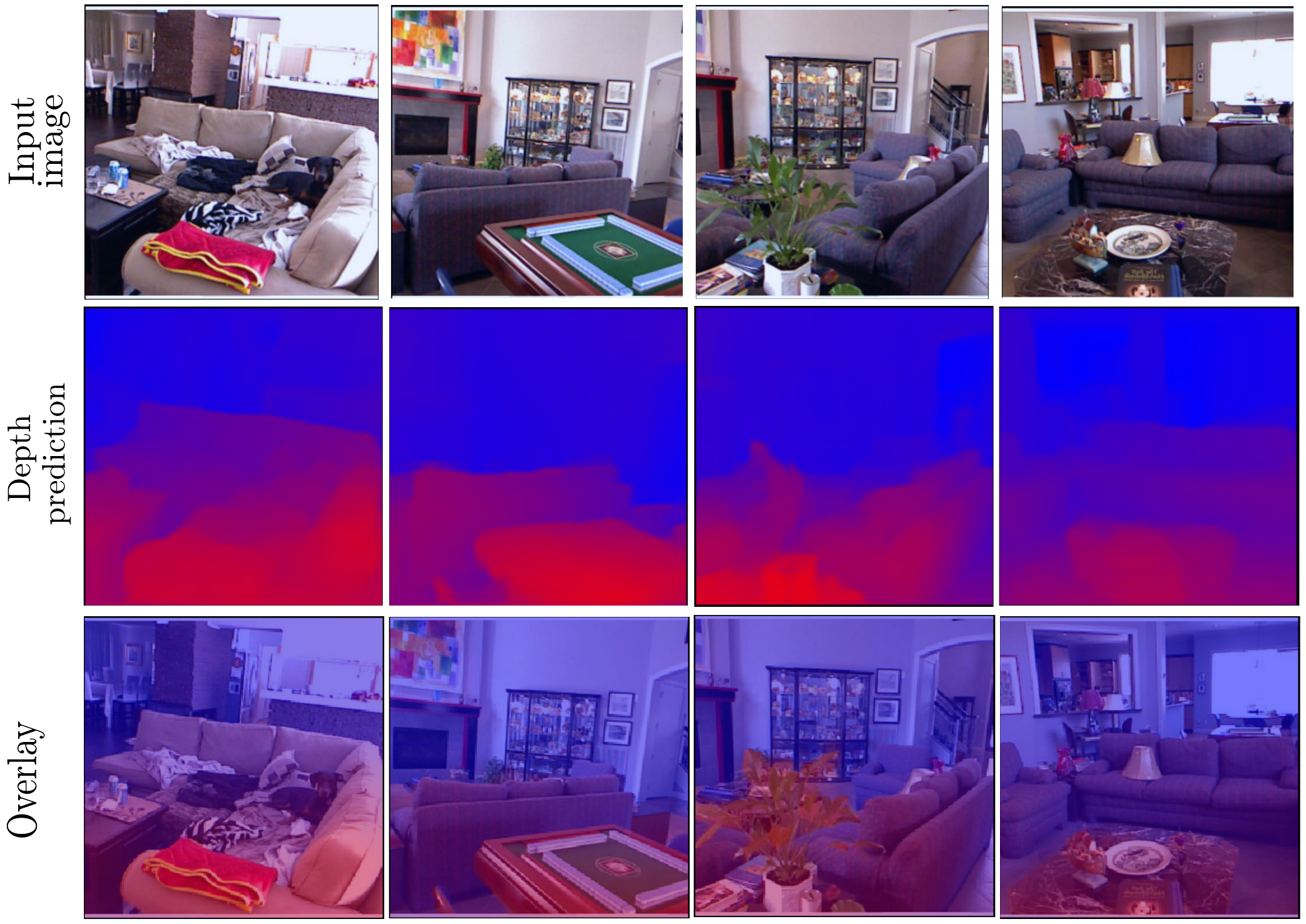}
    \caption{\textbf{Qualitative visualization of depth prediction.} First row: the input image; second row: the depth prediction; third row: the depth prediction overlayed over the input image.}
    \label{fig:depth_visu}
\end{figure}

In Figure~\ref{fig:depth_visu} we display input images from the NYUv2 dataset (validation set),
corresponding depth predictions, alone and overlayed on the input images.

\vspace{5mm}

\vspace{5mm}

\begin{figure}[h!]
\centering
\setlength{\tabcolsep}{1pt}
\newlength{\crocofigwidth}
\setlength{\crocofigwidth}{2.6cm}
\newcommand{\myrotatedlabel}[1]{\rotatebox{90}{\parbox{\crocofigwidth}{\centering \small #1}}}
\resizebox{\linewidth}{!}{
\begin{tabular}{c@{~~~~}cc@{~~~~}cc@{~~~~}cc}
& \multicolumn{2}{c}{\includegraphics[width=\crocofigwidth]{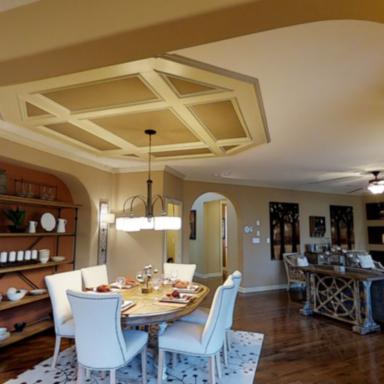}}  
& \multicolumn{2}{c}{\includegraphics[width=\crocofigwidth]{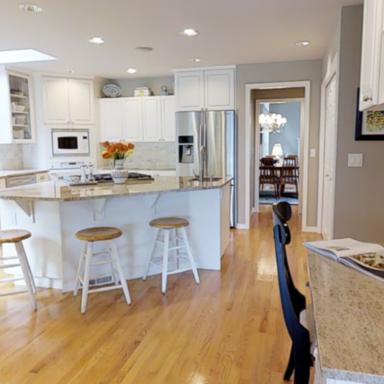}}
& \multicolumn{2}{c}{\includegraphics[width=\crocofigwidth]{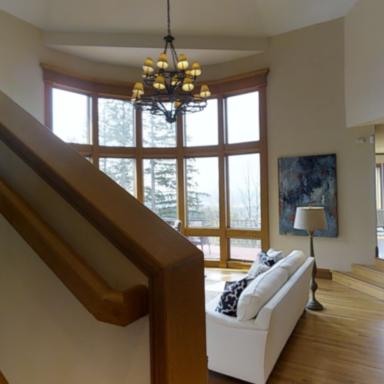}}
\\
& prediction
& ground truth
& prediction
& ground truth
& prediction
& ground truth
\\
\myrotatedlabel{ curvature }
& \includegraphics[width=\crocofigwidth]{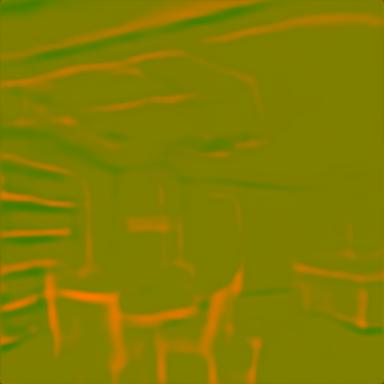}
& \includegraphics[width=\crocofigwidth]{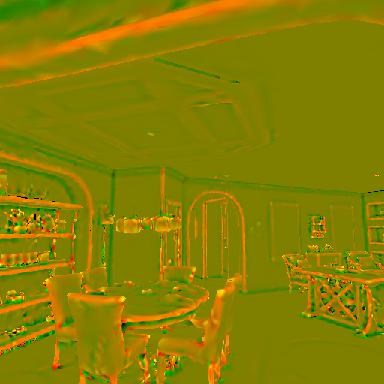}
& \includegraphics[width=\crocofigwidth]{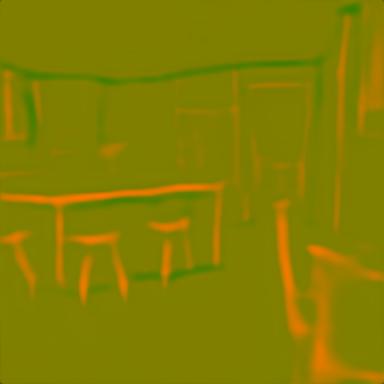}
& \includegraphics[width=\crocofigwidth]{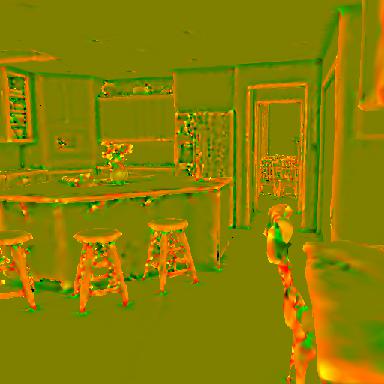}
& \includegraphics[width=\crocofigwidth]{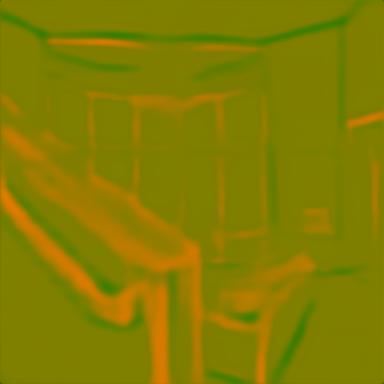}
& \includegraphics[width=\crocofigwidth]{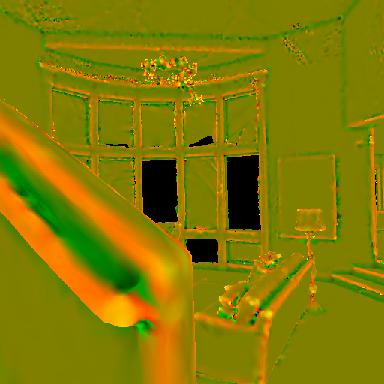}
\\

\myrotatedlabel{ depth }
& \includegraphics[width=\crocofigwidth]{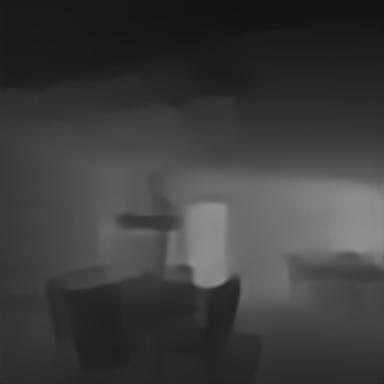}
& \includegraphics[width=\crocofigwidth]{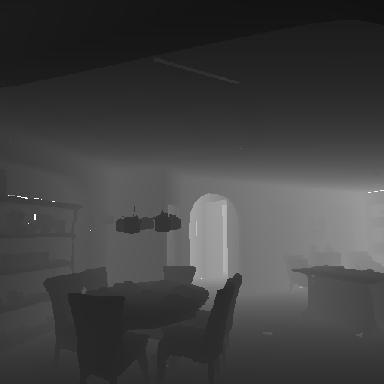}
& \includegraphics[width=\crocofigwidth]{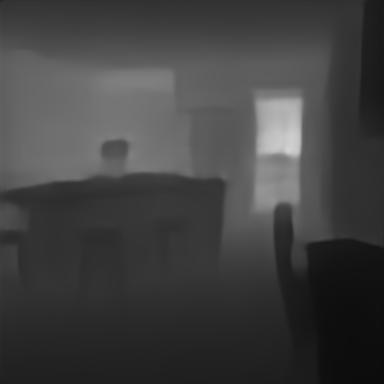}
& \includegraphics[width=\crocofigwidth]{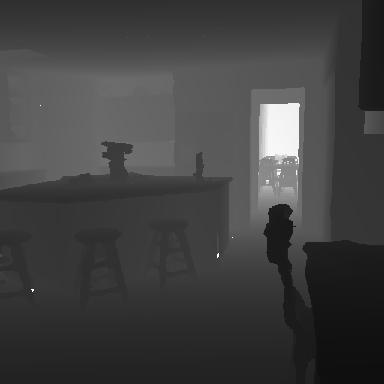}
& \includegraphics[width=\crocofigwidth]{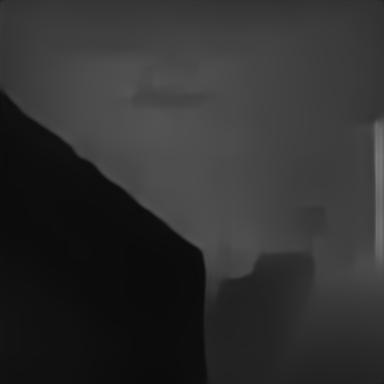}
& \includegraphics[width=\crocofigwidth]{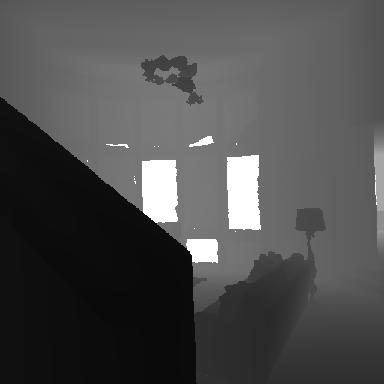}
\\

\myrotatedlabel{ edges }
& \includegraphics[width=\crocofigwidth]{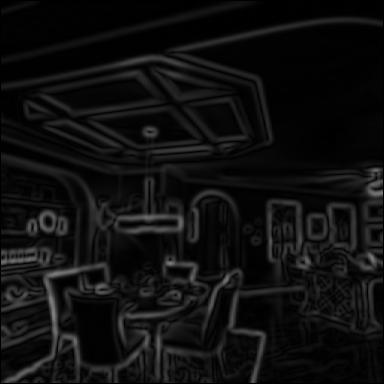}
& \includegraphics[width=\crocofigwidth]{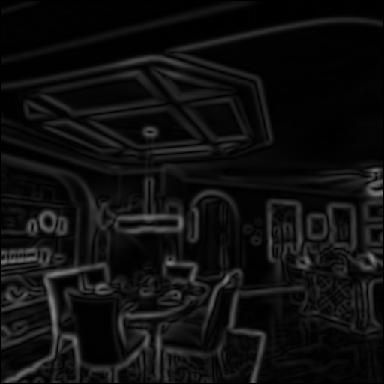}
& \includegraphics[width=\crocofigwidth]{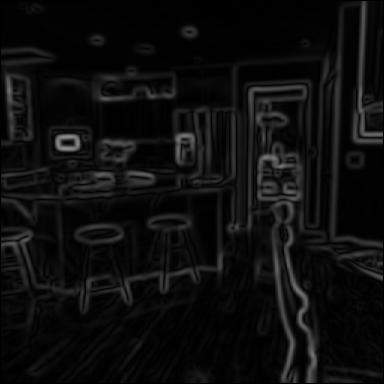}
& \includegraphics[width=\crocofigwidth]{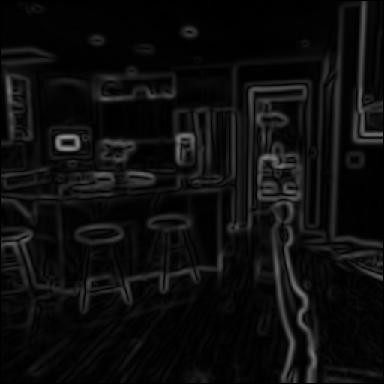}
& \includegraphics[width=\crocofigwidth]{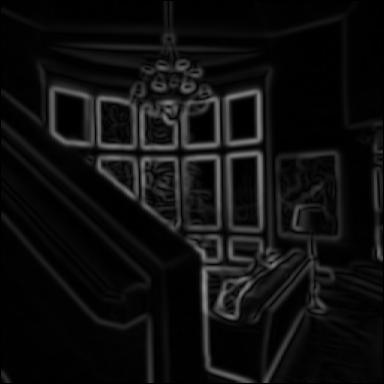}
& \includegraphics[width=\crocofigwidth]{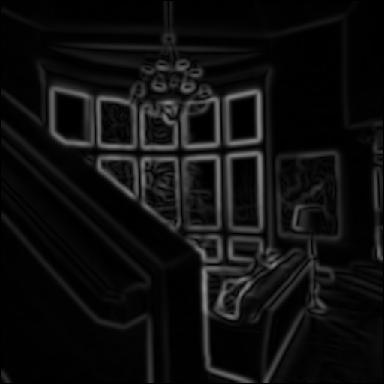}
\\

\myrotatedlabel{ keypoints2d }
& \includegraphics[width=\crocofigwidth]{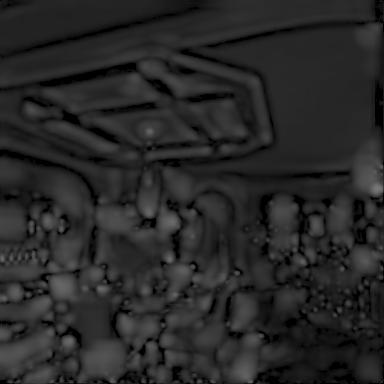}
& \includegraphics[width=\crocofigwidth]{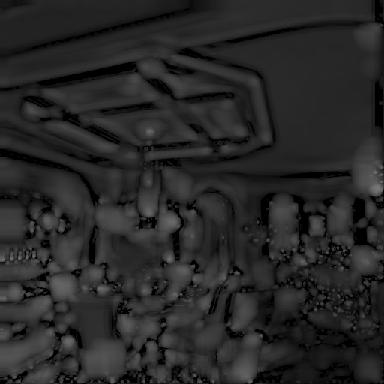}
& \includegraphics[width=\crocofigwidth]{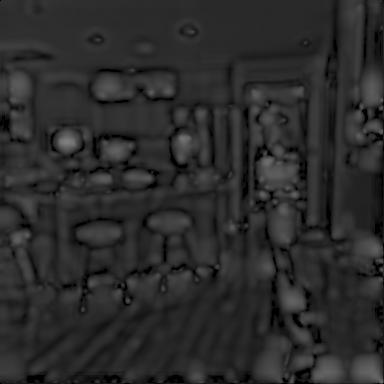}
& \includegraphics[width=\crocofigwidth]{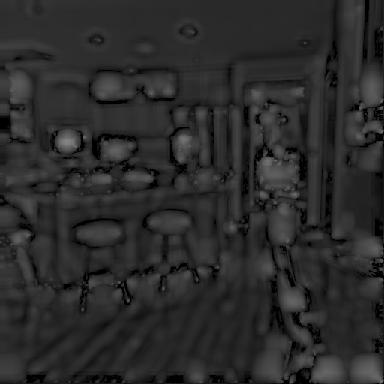}
& \includegraphics[width=\crocofigwidth]{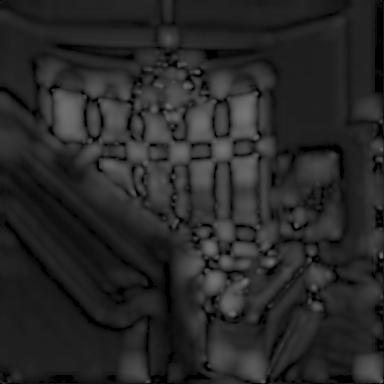}
& \includegraphics[width=\crocofigwidth]{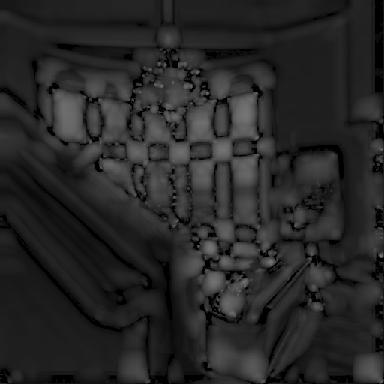}
\\

\myrotatedlabel{ keypoints3d }
& \includegraphics[width=\crocofigwidth]{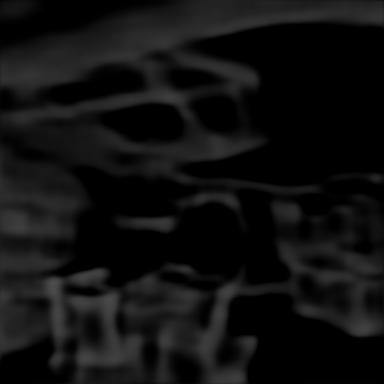}
& \includegraphics[width=\crocofigwidth]{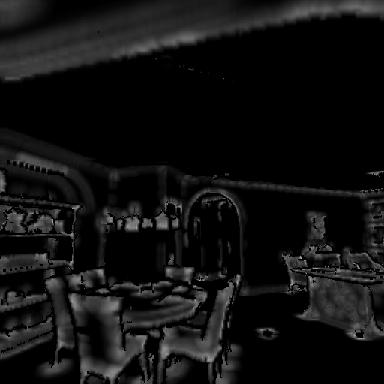}
& \includegraphics[width=\crocofigwidth]{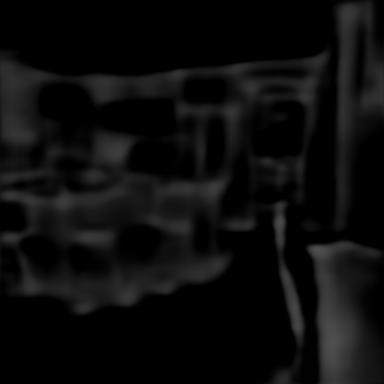}
& \includegraphics[width=\crocofigwidth]{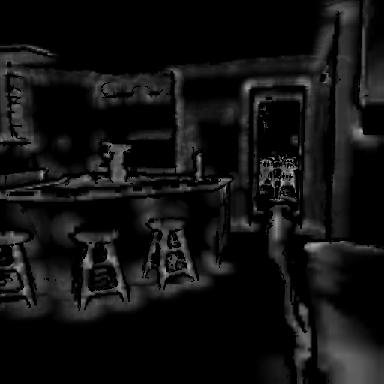}
& \includegraphics[width=\crocofigwidth]{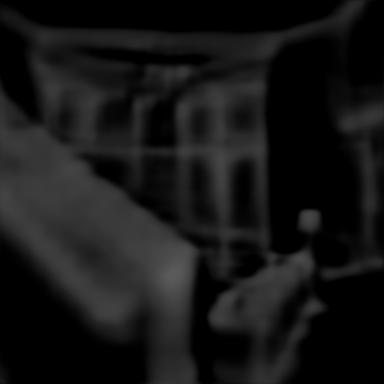}
& \includegraphics[width=\crocofigwidth]{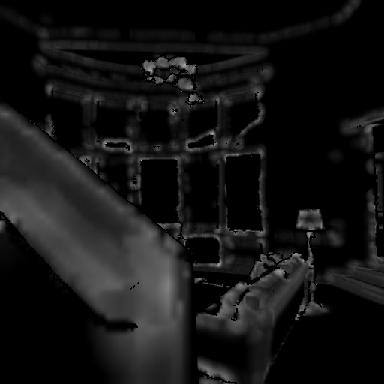}
\\

\myrotatedlabel{ normal }
& \includegraphics[width=\crocofigwidth]{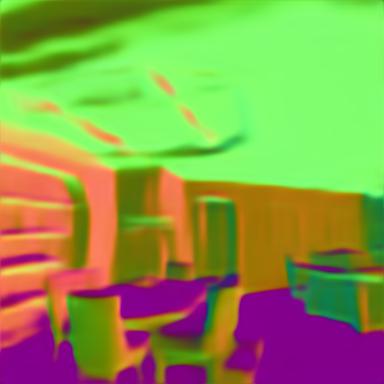}
& \includegraphics[width=\crocofigwidth]{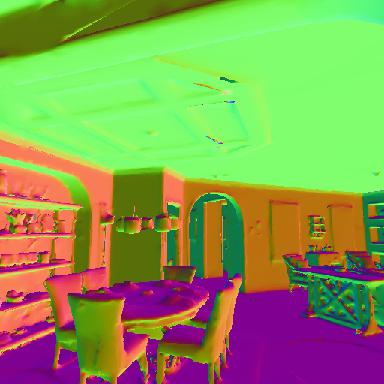}
& \includegraphics[width=\crocofigwidth]{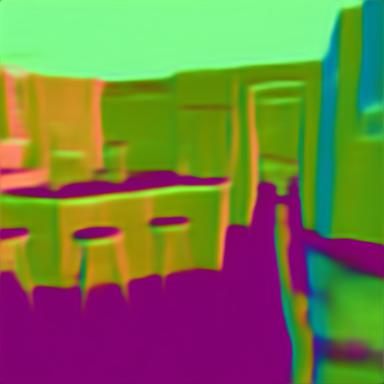}
& \includegraphics[width=\crocofigwidth]{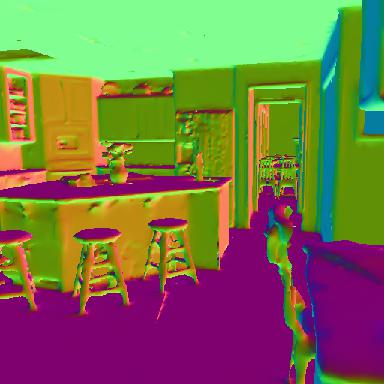}
& \includegraphics[width=\crocofigwidth]{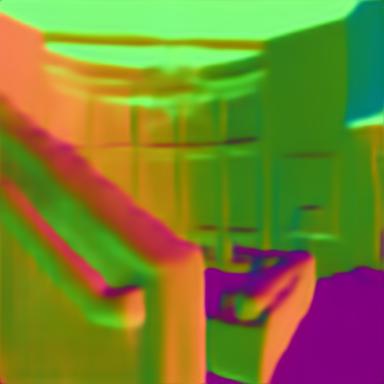}
& \includegraphics[width=\crocofigwidth]{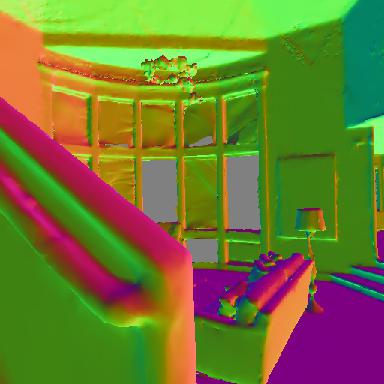}
\\

\myrotatedlabel{ occlusion }
& \includegraphics[width=\crocofigwidth]{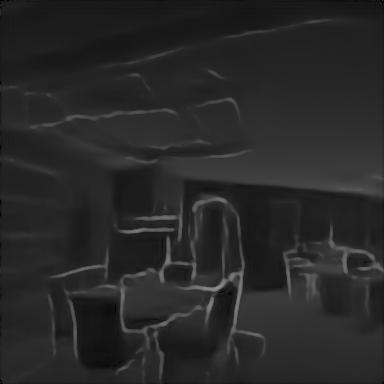}
& \includegraphics[width=\crocofigwidth]{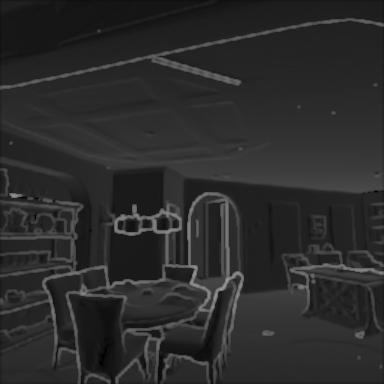}
& \includegraphics[width=\crocofigwidth]{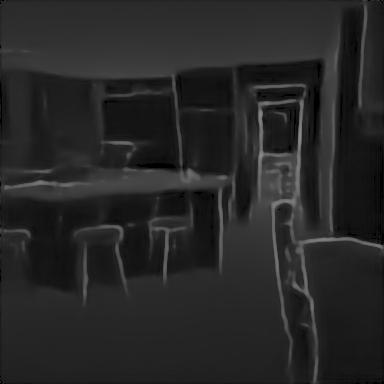}
& \includegraphics[width=\crocofigwidth]{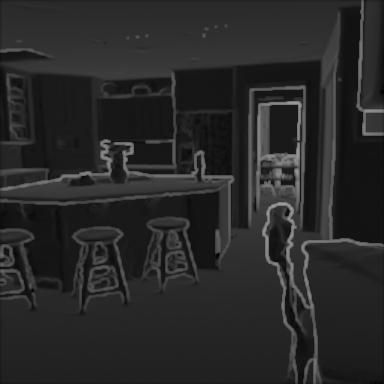}
& \includegraphics[width=\crocofigwidth]{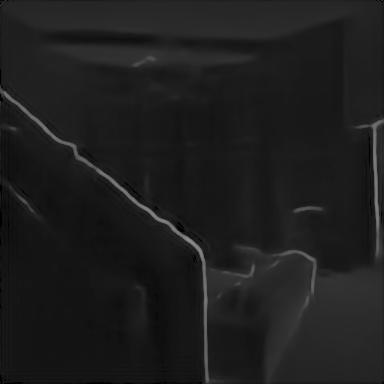}
& \includegraphics[width=\crocofigwidth]{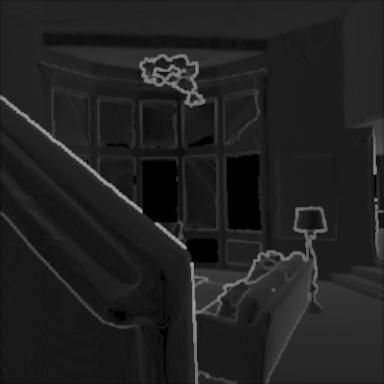}
\\

\myrotatedlabel{ reshading }
& \includegraphics[width=\crocofigwidth]{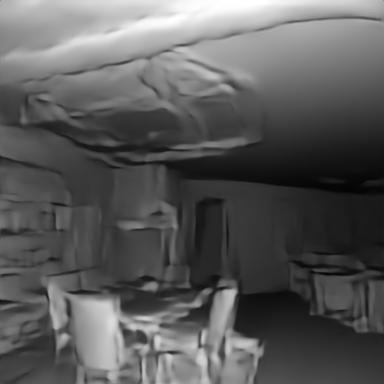}
& \includegraphics[width=\crocofigwidth]{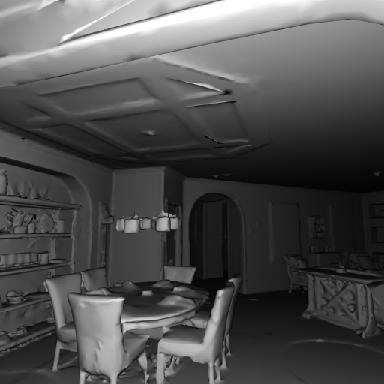}
& \includegraphics[width=\crocofigwidth]{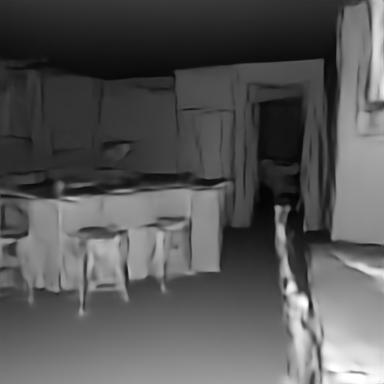}
& \includegraphics[width=\crocofigwidth]{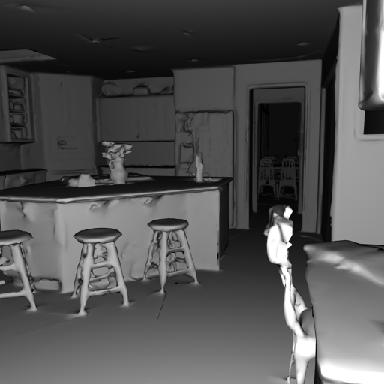}
& \includegraphics[width=\crocofigwidth]{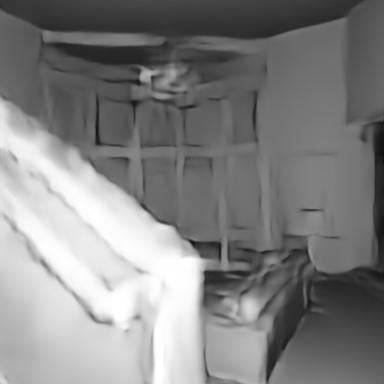}
& \includegraphics[width=\crocofigwidth]{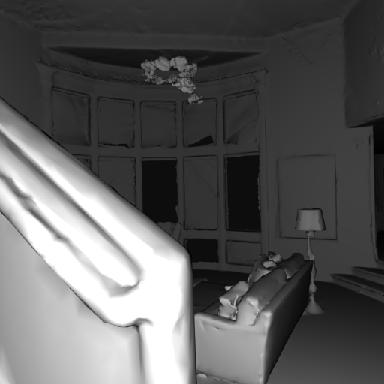}
\\
\end{tabular}
}
\caption{\label{fig:suppmat_taskonomy}\textbf{Example results on Taskonomy.} For the images in the top row, we show the prediction made by our finetuned model and the ground truth, for each of the 8 tasks.}
\end{figure}

Furthermore, in Figure~\ref{fig:suppmat_taskonomy} we present results on the Taskonomy dataset for the 8 dense regression tasks.

\clearpage

\subsection{Application to absolute pose regression from a single image}
\label{supsub:apr}

\noindent \textbf{Task and setting.} We apply our pre-trained model to the task of
monocular absolute pose regression, where a model directly predicts the absolute camera pose 
of a given image in a fixed and given environment for which it was specifically trained.
In contrast to  classical structure-based methods, camera localization methods based on deep neural networks directly regress the absolute pose of a query image~\cite{KendallICCV15PoseNetCameraRelocalization,KendallCVPR17GeometricLossCameraPoseRegression,BrahmbhattCVPR18GeometryAwareLocalization,WangAAAI20AtLocAttentionGuidedCameraLocalization} and do not require database images or 3D maps at test time. 
While these methods are in general significantly less accurate than structure-based methods~\cite{SchonbergerCVPR16StructureFromMotionRevisited,SnavelyIJCV08ModelingTheWorldFromInternetPhotoCollections,HeinlyCVPR15ReconstructingTheWorldSixDays,HumenbergerX20RobustImageRetrievalBasedVisLocKapture},
they only require images and their corresponding camera poses as training data. 

We use the evaluation code and protocol from AtLoc~\cite{WangAAAI20AtLocAttentionGuidedCameraLocalization}, a recent state-of-the-art improvement of the seminal PoseNet~\cite{KendallICCV15PoseNetCameraRelocalization} work, which adds an attention module between the encoder output and the pose regressor heads to reweigh the encoded features. 
The pose regressor is composed of two independent MLP heads: one for predicting the camera pose and one for the camera rotation, represented as a quaternion. 
In our experiments we append the same regression head (attention module and pose regressor heads) on top of the CroCo encoder, with a global average pooling layer inserted in-between to get a global image representation from the encoder tokens. 
To finetune the model,
we rely on the loss proposed in \cite{BrahmbhattCVPR18GeometryAwareLocalization} and also used in \cite{WangAAAI20AtLocAttentionGuidedCameraLocalization}:
\begin{equation}
\Loss=e^{-\beta} \|\bm{p}-\hat{\bm{p}}\|_1 + e^{-\gamma} \|\log \bm{q}-\log \hat{\bm{q}}\|_1 +\beta + \gamma
\end{equation}
where $\beta$ and $\gamma$ are learned weights that balance the position loss and rotation loss and $\log \bm{q}$ is the logarithmic form of a unit quaternion (\ie, the corresponding rotation vector). To remove ambiguity between $\bm{q}$ and $-\bm{q}$, all quaternions are restricted to the same hemisphere. 
The finetuning parameters used for  the pose regression are the followings:

\begin{center}
\small
\begin{tabular}{lc}
\toprule{}
Hyperparameters & Value  \\
\midrule{}
Optimizer & Adam (Base learning rate = 5e-5; weight decay = 5e-4)\\
Training epochs & 500 (with batch size 64) \\
dropout rate probability & 0.5 \\
weights $\beta$ and $\gamma$ & initialized with 0 respectively -3 \\
\midrule
Finetuning dataset & 7-Scenes~\cite{glocker2013_7scenes} \\
Input resolution & $224 \times 224$ (Input field of view  49\td{}) \\
Augmentation &  color jittering and homographies\\
\bottomrule{}
\end{tabular}
\end{center}

We benchmark performance on the 7-Scenes dataset~\cite{glocker2013_7scenes} on the test split of each scene considered independently.
We considered  the  RGB camera pose annotations from  Kapture~\cite{HumenbergerX20RobustImageRetrievalBasedVisLocKapture} and the images were rescaled and cropped to a $224 \times 224$ resolution with a constant 49\td{} field of view (the largest value fitting in the original images).

\PAR{Results}
We evaluate the impact of the choice of backbone and pre-training method in Table~\ref{tab:avgRes} as a function of the size of the training set.
The first row corresponds to a ResNet34 backbone, as used in AtLoc, pre-trained with supervision on ImageNet. Other rows correspond to a ViT-Base/16 backbone with different pre-training methods and datasets (MAE, MultiMAE, CroCo).
We observe that when using the full training set, all models achieve similar performance, except for the model initialized by MultiMAE which consistently underperforms. Note that CroCo achieves peak validation performance in approx. $50-100$ epochs, while ResNet34 pre-trained models need approximately $500$ epochs to converge.
To better assess the performance of the pre-training model, 
we significantly reduce the size of the training set and report results in Table~\ref{tab:avgRes} using only 5, 10 and 20\% of the full training set. 
We observe that the original AtLoc model, a pre-trained ResNet34 encoder, is unable to learn from such a small amount of data even with data augmentations (random color jittering and homographies simulating camera rotations). Models pre-trained using MAE/MultiMAE perform better, but Croco pre-training leads to the best performance.
Finally, while a model trained from scratch (two last rows in Table~\ref{tab:avgRes}) can achieve decent localization performances using the full training set compared to the other baselines, it generalizes poorly when trained using a more limited amount of data.\\

\begin{table}[h]
\small
\caption{\textbf{Absolute pose regression results} as averaged median errors over the 7 scenes for different backbones, pre-training methods and datasets. Each column corresponds to different ratios of the training set. Best result per column on bold and second best underlined.} 
\label{tab:avgRes}
\setlength{\tabcolsep}{3pt}
\begin{tabular}{llccccc}
\toprule{}
Architecture  & pre-training                & 100\%                      & 20\%                       & 10\%                       & 5\%  			             \\
\midrule{}%
ResNet34 & Supervised (ImageNet)              & {\bf 27.1cm}, 9.0\td{}             & 34.0cm, 10.9\td{}            & 43.7cm, 130.\td{}            & 62.3cm, 17.3\td{}          \\
ViT-Base/16 & MAE (ImageNet)                   & 27.9cm, 9.0\td{}             & \ul{28.0cm}, 8.5\td{}             & \ul{30.6cm}, 9.2\td{}           & \ul{34.4cm}, \bf{9.4}\td{}      \\
ViT-Base/16 & MAE (Habitat)                    & 28.3cm, 9.1\td{}           & \ul{28.0cm}, \ul{8.3\td}{}             & 30.7cm, \ul{9.1}\td{}      & 35.3cm, 10.1\td{}          \\
 ViT-Base/16 & MultiMAE (ImageNet)              & 33.1cm, 9.8\td{}           & 32.6cm, 9.7\td{}           & 36.8cm, 10.8\td{}          & 44.1cm, 12.2\td{}          \\
 ViT-Base/16 & {\color{OliveGreen}\bf{CroCo}} CrossBlock (Habitat)
& \ul{27.7cm}, \bf{8.6}\td{} & \bf{26.3cm}, \bf{7.3}\td{} & \bf{29.0cm}, \bf{8.3}\td{}   & {\bf 32.7cm}, \ul{9.5}\td{} \\ 
\midrule
ResNet34 & Random &   28.1cm, \ul{8.9}\td{} &  36.6cm, 11.4\td{} & 51.3cm, 14.0\td{} & 69.3cm, 17.6\td{} \\
 ViT-Base/16 & Random & 29.1cm, 9.3\td{} & 38.3cm, 11.6\td{} & 43.6cm, 12.5\td{} & 52.7cm, 14.1\td{} \\
\bottomrule{}
\end{tabular}
\end{table}

\section{Further details and results on the binocular downstream tasks}
\label{sup:binocular}

\subsection{Optical Flow estimation}
\label{supsub:flow}

\PAR{Experimental details}
We treat optical flow as a straightforward regression task and do not change the pre-training architecture except for modifying the regression head to output two flow channels instead of 3 RGB color channels: the two images are input as such in the Siamese encoders, and the decoder regresses two flow values $(u, v)$ for each pixel using a simple linear head running independently for each output token.
We use a \textit{CrossBlock} decoder with 8 layers (similar results are obtained with a \textit{CatBlock} decoder). 
Training details are provided in Table~\ref{tab:hyperparams}.\\

\begin{table}[b]
\caption{\label{tab:hyperparams}
\textbf{Parameter settings for optical flow, relative pose regression and stereo matching.}}
\centering
\small
\begin{tabular}{lccc}
\toprule{}
Hyperparameters & Optical Flow & Relative Pose Regression & Stereo Matching  \\
\midrule{}
Optimizer            & AdamW~\cite{LoshchilovICLR19AdamW}
                                  & AdamW~\cite{LoshchilovICLR19AdamW}                              &AdamW~\cite{LoshchilovICLR19AdamW}\\
Base learning rate   & 3e-5       & 1e-7           &1e-4\\
Weight Decay         & 0.05       & -              & - \\
Batch size           & 20         & 64             & 8 \\
Adam $\beta$         &(0.9, 0.95) & (0.9, 0.95)    & (0.9, 0.95) \\
Learning rate sched. & Cosine decay & Cosine decay & 1 cycle\\
Finetuning epochs    & 100        & 30             & 400 \\
Linear warmup epochs & 1          & 1              & 1\\
Translation weight $\lambda$ & -  & 100m   & - \\
\midrule{}
Finetuning dataset   & AutoFlow~\cite{autoflow}
                                 & 7-scenes~\cite{glocker2013_7scenes} 
                                        &VKITTI~\cite{cabon2020virtualkitti} \\
Input resolution & $224 \times 224$ & $224 \times 224$  & $1242 \times 375$ \\
Augmentation        & random crop,  & homography, & random horizontal flip, \\
                    & color jitter & color jitter & color jitter\\
Input field of view &  -         & 49\td{}       & - \\
\bottomrule
\end{tabular}
\end{table}

\begin{figure}[h]
    \centering
    \includegraphics[width=0.9\textwidth]{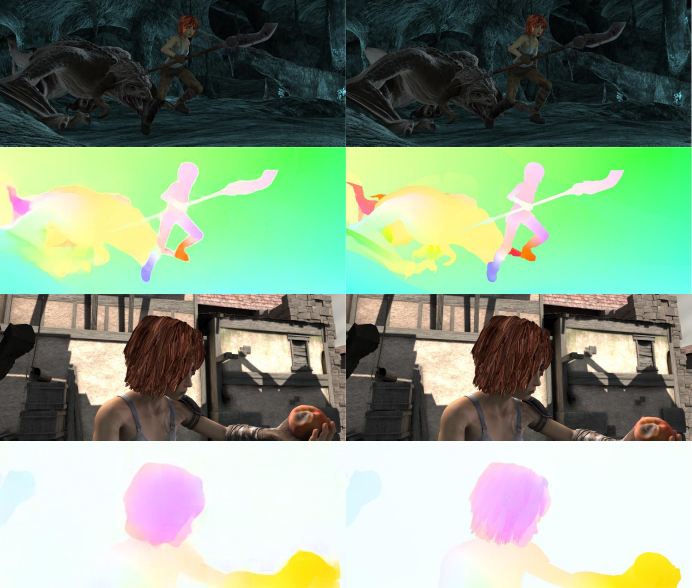}
     \caption{\textbf{Qualitative visualization of the estimated flow} on some examples from the MPI-Sintel training set, unseen during training. First and third rows: input image pair, second and fourth row: predicted (\textit{left}) and ground-truth (\textit{right}) flows.}
    \label{fig:flowSINTEL}
\end{figure}

\PAR{Qualitative Results}
We show in Figure~\ref{fig:flowSINTEL} qualitative visualization of the flow predicted by our method.
Overall, we observe that the flow is correctly estimated, even in the case of varied and fast motion. As a limitation, the estimation gets slightly inaccurate when occlusions happen, and blurry on extremely fine-grained motion such as hair tips, as for most methods. Again, we wish to emphasize that the model was  finetuned only on 40,000 images without any sort of elaborated data augmentation.

\subsection{Relative pose regression}

\PAR{Model Architecture}
We replace the original head of the decoder pre-trained with CroCo by a simple head regressing a relative pose $(\bm{R}, \bm{t}) \in SO(3) \times \mathbb{R}^3$. This head consists of the following layers. First, a linear projection reduces the dimension of tokens produced by the decoder to 64, in order to limit the computation costs. Second, these tokens are flattened into a unique feature vector, which is processed by a MLP (with a hidden layer of size 1024 and ReLU activations) to regress a 12D output. It is reshaped into an affine transformation $(\bm{M}, \bm{t}) \in \mathbb{R}^{3 \times 3} \times \mathbb{R}^{3}$. Lastly we use a differentiable special Procrustes layer~\cite{bregier2021deepregression} to orthonormalize $\bm{M}$ into a rotation matrix $\bm{R}$ and to predict the relative pose $(\bm{R}, \bm{t})$.

\PAR{Experimental details}
The RGB camera pose annotations for the 7-scenes dataset~\cite{glocker2013_7scenes} are obtained from Kapture~\cite{HumenbergerX20RobustImageRetrievalBasedVisLocKapture}.
To select training pairs, we extract global AP-GeM-18 descriptors~\cite{revaud2019apgem} for every image, and match each training image with its 20 closest neighbors according to their descriptor similarity. Our training set is composed of 26k images and 520k training pairs in total.
We rescale and crop the images to a square $224 \times 224$ resolution with a constant 49° field of view (the largest value fitting in the original images), and we augment the training image pairs with random permutations, color jittering and homographies that simulate camera rotation.

In our experiments, we trained models using the AdamW~\cite{LoshchilovICLR19AdamW} optimizer using settings described in Table~\ref{tab:hyperparams}.
We found the training to be quite sensitive to its initialization, and we had to try multiple random seeds to achieve decent performances on the training set when training the decoder from scratch with an encoder pre-trained with MAE on Habitat. Using an encoder pre-trained with MAE on ImageNet furthermore always led to poor performance on both the training and test sets in our experiments. We had none of these issues however when using an encoder and decoder pre-trained using CroCo.

\subsection{Stereo image matching}
\label{ssec:stereo}

\PAR{Task and settings}
In this section, we experiment on the binocular downstream task of stereo matching~\cite{poggi2021synergies}. 
We finetune a pre-trained CroCo model to predict pixel-wise disparity values, by simply replacing the linear prediction head. 
The model takes two rectified images as input and predicts the disparity of every pixel by matching corresponding pixels in the images. We use a MSE-log loss for finetuning and report the error using the 3-pixel 5\% discrepancy~\cite{chang2018pyramid,cheng20leastereo}.
We evaluate the stereo matching downstream task on the Virtual KITTI dataset~\cite{cabon2020virtualkitti} which consists of 5 synthetic sequences cloned from the KITTI tracking benchmark~\cite{kitti}. The dataset additionally provides 9 variants of these 5 sequences under different weather conditions (\eg `fog', `rain', \etc) and modified camera configurations (\eg rotation to left or right by 15 or 30 degrees). 
We downscale VKITTI images 
to a resolution of 224 $\times$ 742, preserving the aspect ratio and crop to 224 $\times$ 736. Additional details on the finetuning hyper-parameters and settings are described in Table~\ref{tab:hyperparams}.

\begin{figure}[ttt]
    \centering
    \includegraphics[width=\columnwidth]{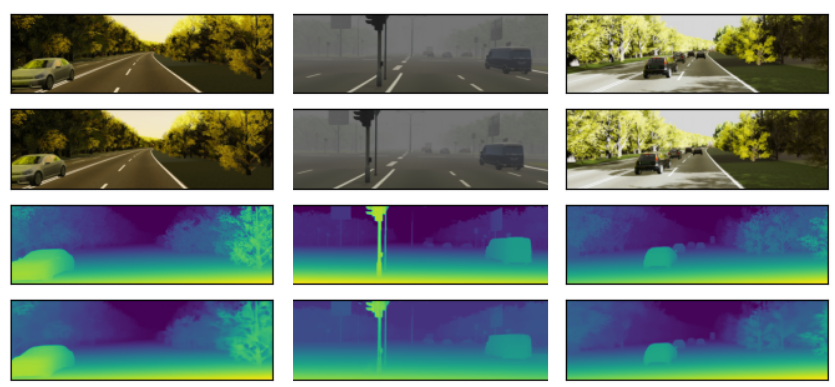} \hfill
    \caption{\textbf{Qualitative visualization of the estimated disparity} on examples from the VKITTI validation set. First row: input left image; second row: input right image, third row: ground-truth (left) disparity, forth row: predicted (left) disparity. The three examples come from ``sunset''  (left column), ``fog'' (center) and ``clone'' (right column) weather conditions.}
    \label{fig:disparity_vkitti}
\end{figure}

\PAR{Results}
In Table~\ref{tab:stereo-vkitti} we report results obtained on all 10 weather condition and camera orientation variants in the VKITTI dataset. We compare results obtained when finetuning the CroCo model and the MAE model with a randomly initialized decoder, pre-trained on Habitat.
We observe that CroCo pre-training leads to significantly better results for all scenarios. Next, we compare to the state-of-the-art methods for stereo matching, in particular, PSMNet~\cite{chang2018pyramid}, LaC-GweNet~\cite{liu22localsimilarity} and LEAStereo~\cite{cheng20leastereo}.
Finetuning a model pre-trained with CroCo achieves results competitive with the state of the art, without any task-specific model design, such as spatial pyramid pooling~\cite{chang2018pyramid}, hierarchical neural search in 4D feature volume~\cite{cheng20leastereo} or local similarity patterns for explicit neighbor relationships~\cite{liu22localsimilarity}. 
We point out that, in contrast to the other methods, our model is directly finetuned on VKITTI without pre-training on the large SceneFlow dataset~\cite{flyingthings}. 
In Figure~\ref{fig:disparity_vkitti} we visualize the disparity prediction by our method, for three different conditions in the VKITTI dataset. 

\input{plots/stereo-vkitti}

\clearpage

\section{Compute resources, code and dataset assets}
\label{sup:asset}

\subsection{Floating point operations (FLOPs)}

We calculate the number of floating point operations (or FLOPs) in the most commonly accepted manner, \ie, counting additions and multiplications as separate floating-point operations.
To that aim, we borrow and slightly adapt the \texttt{flops\_computation.py} code from~\cite{clark2020electra}. 
Note that most of the FLOPs comes from matrix-matrix multiplication operations in the attention and feed-forward layers. For a standard ViT (encoder only), the number of FLOPs can thus be approximated as
\begin{equation*}
\#FLOPs \simeq 2L\left(F_{attn-kqv}+F_{attn-scores}+F_{attn-avg}+F_{attn-out}+F_{ff-in}+F_{ff-out}\right)
\end{equation*}
where
\begin{eqnarray*}
F_{attn-kqv} & = & 3ND^{2}\\
F_{attn-scores} & = & DN^2\\
F_{attn-avg} & = & DN^2\\
F_{attn-out} & = & ND^{2}\\
F_{ff-in} & = & 4ND^{2}\\
F_{ff-out} & = & 4ND^{2}.
\end{eqnarray*}

For instance, for ViT-Base/16, setting the number of blocks $L=12$, the number of tokens $N=14^2$ and the embedding dimension $D=768$, we find 34.7 GFLOPs with the formula above, which is close to the 35.3 GFLOPs calculated including all other operations (layer norms, patch embeddings, \etc). 
Note that we report exact theoretical FLOPs (counting all operations) in the paper.

\subsection{Compute resources}

Pre-training a CroCo model for $400$ epochs takes about 64 GPU-days on NVIDIA V100. For instance on a 4-GPU server, this is about two weeks.

\subsection{Assets used in this submission}

We provide below 
an overview of assets used in our experiments and their licenses.

\begin{center}
\resizebox{\linewidth}{!}{
\begin{tabular}{p{0mm}ll}
\toprule
\multicolumn{2}{l}{Asset} & License \\
\midrule
\multicolumn{3}{l}{\textbf{Habitat pre-training}} \\
& HM3D~\cite{ramakrishnan2021hm3d} &  academic, non-commercial research \href{https://matterport.com/matterport-end-user-license-agreement-academic-use-model-data}{[hyperlink]}\\
& ScanNet~\cite{dai2017scannet} & non-commercial research and education \href{http://kaldir.vc.in.tum.de/scannet/ScanNet_TOS.pdf}{[hyperlink]}
\\
& Replica~\cite{replica19arxiv} & non-commercial research and education \href{https://github.com/facebookresearch/Replica-Dataset/blob/main/LICENSE}{[hyperlink]}
\\
& ReplicaCAD~\cite{szot2021habitat} & Creative Commons Attribution 4.0 International (CC BY 4.0) \href{https://aihabitat.org/datasets/replica_cad/}{[hyperlink]} \\
& Habitat simulator~\cite{habitat19iccv} & MIT \href{https://github.com/facebookresearch/habitat-sim/blob/main/LICENSE}{[hyperlink]} 
\\
\midrule
\multicolumn{3}{l}{\textbf{High-level semantic tasks}} \\
& ImageNet-1K~\cite{ILSVRC15imagenet} & non-commercial research and education \href{https://www.image-net.org/download.php}{[hyperlink]} \\
& ADE~\cite{ade} & images: non-commercial research and education -- annotations: Creative Commons BSD-3 \href{https://groups.csail.mit.edu/vision/datasets/ADE20K/terms/}{[hyperlink]}\\
\midrule
\multicolumn{3}{l}{\textbf{Monocular 3D vision tasks}} \\
& NYUv2~\cite{nyuv2} & public \href{https://cs.nyu.edu/~silberman/datasets/nyu_depth_v2.html}{[hyperlink]} 
\\
& Taskonomy~\cite{zamir2018taskonomy} & non-commercial research and education \href{https://github.com/StanfordVL/taskonomy/blob/master/data/LICENSE}{[hyperlink]} 
\\
\midrule
\multicolumn{3}{l}{\textbf{Optical flow}} \\
& AutoFlow~\cite{autoflow} & Create Commons Attribution 4.0 International (CC BY 4.0) \href{https://autoflow-google.github.io/\#data}{[hyperlink]}
\\
& MPI-Sintel~\cite{sintel} & images: Creative Commons Attribution 3.0 (CC BY 3.0) -- copyright Blender Foundation | www.sintel.org \href{http://sintel.is.tue.mpg.de/}{[hyperlink]}
\\
\midrule
\multicolumn{3}{l}{\textbf{Absolute / relative pose regression}} \\
& 7-scenes~\cite{glocker2013_7scenes} & non-commercial \href{https://www.microsoft.com/en-us/research/wp-content/uploads/2016/02/7-scenes-msr-la-dataset-7-scenes.rtf}{[hyperlink]} \\
& Kapture package~\cite{HumenbergerX20RobustImageRetrievalBasedVisLocKapture} & BSD 3-Clause Revised \href{https://github.com/naver/kapture/blob/main/LICENSE}{[hyperlink]}\\
\midrule
\multicolumn{3}{l}{\textbf{Stereo matching}} \\
& Virtual KITTI~\cite{cabon2020virtualkitti} & Creative Commons Attribution-NonCommercial-ShareAlike 3.0 \href{https://europe.naverlabs.com/research/computer-vision/proxy-virtual-worlds-vkitti-1/}{[hyperlink]} \\
\midrule
\multicolumn{3}{l}{\textbf{Code and pre-trained models}} \\
& MAE~\cite{he2021mae} & Creative Commons Attribution-NonCommercial 4.0 International (CC BY NC 4.0) \href{https://github.com/facebookresearch/mae/blob/main/LICENSE}{[hyperlink]}
\\
& MultiMAE~\cite{bachmann2022multimae} & Creative Commons Attribution-NonCommercial 4.0 International (CC BY NC 4.0) \href{https://github.com/EPFL-VILAB/MultiMAE/blob/main/LICENSE}{[hyperlink]} \\
& DINO~\cite{CaronICCV21DINO} & Apache License 2.0 \href{https://github.com/facebookresearch/dino/blob/main/LICENSE}{[hyperlink]}\\
\bottomrule
\end{tabular}
}
\end{center}

\clearpage 

\section{Further visual examples}
\label{sup:visualex}

\begin{figure}[h]
\centering
\small
\setlength{\tabcolsep}{1pt}
\setlength{\crocofigwidth}{2.6cm}
\newcommand{\myrotatedlabel}[1]{\rotatebox{90}{\parbox{\crocofigwidth}{\centering \small #1}}}
\begin{tabular}{cccccc}
\myrotatedlabel{ Reference input}
& \includegraphics[width=\crocofigwidth]{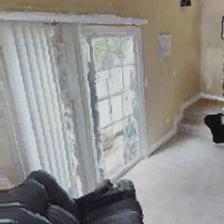}
& \includegraphics[width=\crocofigwidth]{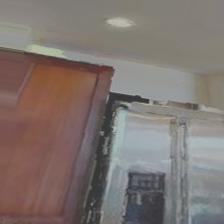}
& \includegraphics[width=\crocofigwidth]{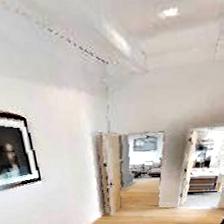}
& \includegraphics[width=\crocofigwidth]{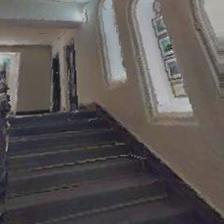}
& \includegraphics[width=\crocofigwidth]{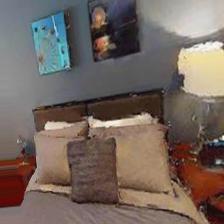}
\\

\myrotatedlabel{Masked input}
& \includegraphics[width=\crocofigwidth]{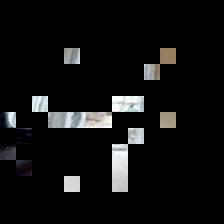}
& \includegraphics[width=\crocofigwidth]{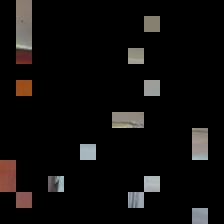}
& \includegraphics[width=\crocofigwidth]{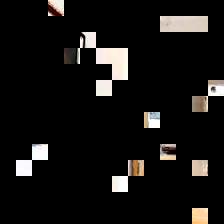}
& \includegraphics[width=\crocofigwidth]{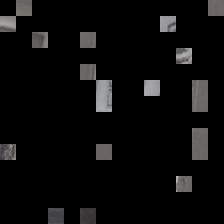}
& \includegraphics[width=\crocofigwidth]{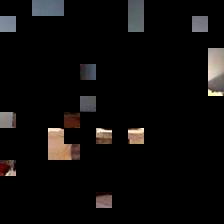}
\\

\myrotatedlabel{Target}
& \includegraphics[width=\crocofigwidth]{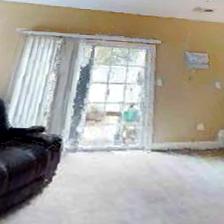}
& \includegraphics[width=\crocofigwidth]{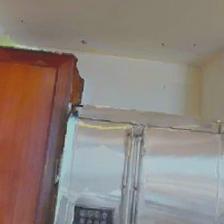}
& \includegraphics[width=\crocofigwidth]{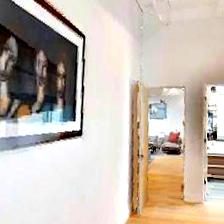}
& \includegraphics[width=\crocofigwidth]{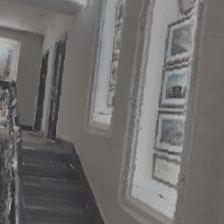}
& \includegraphics[width=\crocofigwidth]{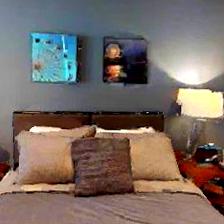}
\\
\midrule{}
\myrotatedlabel{CroCo RGB}
& \includegraphics[width=\crocofigwidth]{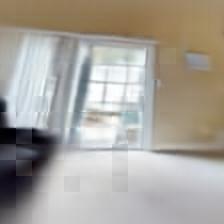}
& \includegraphics[width=\crocofigwidth]{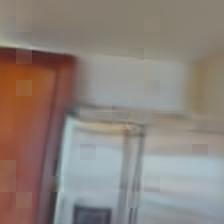}
& \includegraphics[width=\crocofigwidth]{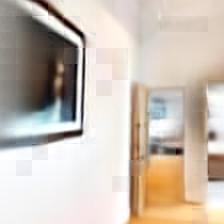}
& \includegraphics[width=\crocofigwidth]{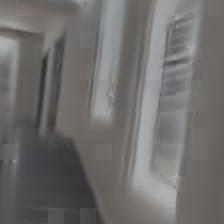}
& \includegraphics[width=\crocofigwidth]{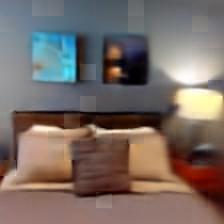}
\\

\myrotatedlabel{CroCo RGB \\ random noise ref.}
& \includegraphics[width=\crocofigwidth]{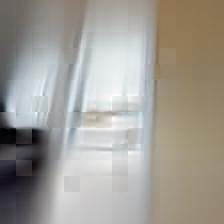}
& \includegraphics[width=\crocofigwidth]{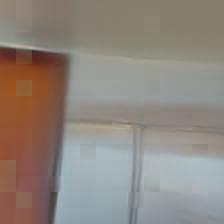}
& \includegraphics[width=\crocofigwidth]{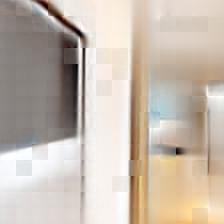}
& \includegraphics[width=\crocofigwidth]{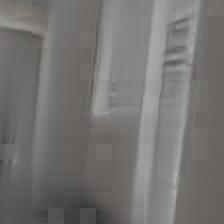}
& \includegraphics[width=\crocofigwidth]{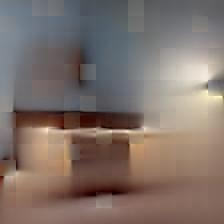}
\\

\midrule{}
\myrotatedlabel{MAE norm}
& \includegraphics[width=\crocofigwidth]{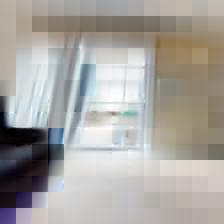}
& \includegraphics[width=\crocofigwidth]{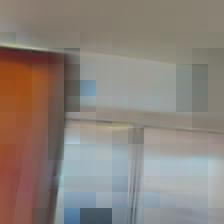}
& \includegraphics[width=\crocofigwidth]{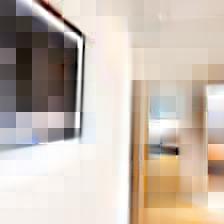}
& \includegraphics[width=\crocofigwidth]{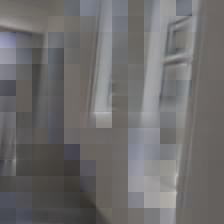}
& \includegraphics[width=\crocofigwidth]{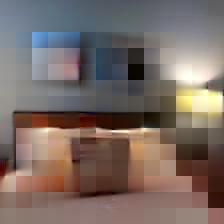}
\\

\myrotatedlabel{CroCo norm}
& \includegraphics[width=\crocofigwidth]{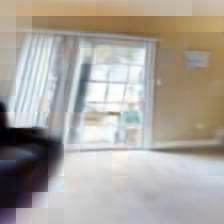}
& \includegraphics[width=\crocofigwidth]{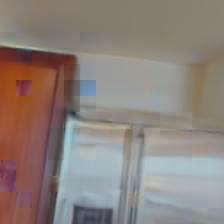}
& \includegraphics[width=\crocofigwidth]{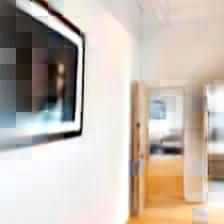}
& \includegraphics[width=\crocofigwidth]{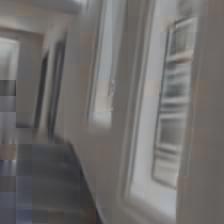}
& \includegraphics[width=\crocofigwidth]{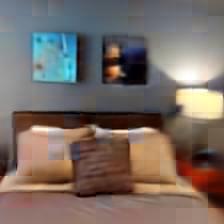}
\\
\end{tabular}
\caption{\label{fig:suppmat_cross_reconstruction}\textbf{Additional reconstruction examples} for scenes unseen during training. We compare image reconstructions obtained using our CroCo model pre-trained with an RGB loss using the reference input image (\emph{CroCo RGB}) or replacing it by random noise (\emph{CroCo RGB random noise ref.}).
We also compare reconstructions of CroCo and MAE after a pre-training on Habitat to reconstruct normalized patches (\emph{CroCo norm} and \emph{MAE norm}). Patch un-normalization is performed using patch statistics from the target image, for visualization purposes.}
\end{figure}

We provide additional reconstruction examples in Figure \ref{fig:suppmat_cross_reconstruction}, where we compare reconstructions obtained using our CroCo model pre-trained with an RGB loss and a reference input image (\emph{CroCo RGB}) with reconstructions obtained after replacing this reference image by an image of random uniform noise, independent at each pixel (\emph{CroCo RGB random noise ref.}).
Cross-view completion enables a decent reconstruction of the masked image in general, except for areas non-visible in the reference image (\eg \emph{CroCo RGB}, left part of the painting in the third column). When replacing the reference image by some random noise, the reconstruction problem becomes similar to inpainting and some parts of the scene may be wrongly reconstructed due to the lack of available information  (\eg wrong size for the window in the first column, missing ice maker in the fridge in the second column).

We also compare reconstructions obtained with CroCo and MAE models pre-trained on Habitat to reconstruct normalized patches (\emph{CroCo norm} and \emph{MAE norm}, last two rows). We use patch statistics from the target image to un-normalize these reconstructions for visualization purposes, which explains that the mean color of each reconstructed patch is relatively well reconstructed in all cases. Yet, we observe that CroCo produces more detailed reconstructions than MAE, the latter often appearing quite tessellated due to an inconsistent reconstruction of normalized patches.

\end{document}

%% file: plots/plotabbrev.tex
\usepackage{pgfplots, pgfplotstable}
\pgfplotsset{compat=newest}
\usepgfplotslibrary{fillbetween}

\newcommand{\oxf}{MidnightBlue}

\newcommand{\paris}{RedViolet}

\newcommand{\valc}{PineGreen}

\newcommand{\ourscolor}{Maroon}
\newcommand{\howcolor}{Black}

\newcommand{\oxfmark}{o}
\newcommand{\parmark}{triangle}
\newcommand{\valmark}{square}

\newcommand{\bone}{CornflowerBlue}
\newcommand{\btwo}{MidnightBlue}
\newcommand{\bthree}{TealBlue}
\newcommand{\bfour}{Blue}

\tikzset{every mark/.append style={solid}}
\pgfplotsset{
	grid=both, width=\linewidth, try min ticks=5,
    legend cell align=left, 
    legend style={fill opacity=0.8},
	ylabel near ticks,
    xlabel near ticks,
    every tick label/.append style={font=\footnotesize},
}

\pgfplotsset{
    oxmed/.style={thick, color=\oxf, mark=o},
    oxhard/.style={thick, dashed, color=\oxf, mark=o},
    pamed/.style={thick, color=\paris, mark=star}, 
    pahard/.style={thick, dashed, color=\paris, mark=star},
    val/.style={thick, color=\valc, mark=x},
    oursoxf/.style={thick, color=\ourscolor, mark=\oxfmark},
    howoxf/.style={thick, dashed, color=\howcolor, mark=\oxfmark},
    ourspar/.style={thick, color=\ourscolor, mark=\parmark},
    howpar/.style={thick, dashed, color=\howcolor, mark=\parmark},
    oursval/.style={thick, color=\ourscolor, mark=\valmark},
    howval/.style={thick, dashed, color=\howcolor, mark=\valmark},
    numean/.style={thick, color=\ourscolor, mark=none},
    numin/.style={thick, color=gray, mark=none},
    wid/.style={thick, color=\ourscolor, mark=o, mark size=0.5pt},
    woid/.style={thick, densely dashdotted, color=RedOrange, mark=o, mark size=0.5pt},
    d2k1/.style={thick, color=CornflowerBlue, mark=\oxfmark},
    d2k2/.style={thick, color=MidnightBlue, mark=\oxfmark},
    d2k4/.style={thick, color=TealBlue, mark=\oxfmark},
    d2k8/.style={thick, color=Blue, mark=\oxfmark},
    d4c256/.style={thick, color=\ourscolor, mark=\oxfmark},
    d8c256/.style={thick, color=Green, mark=\oxfmark},
    b1c/.style={thick, color=\bone, mark=o},
    b2c/.style={thick, color=\btwo, mark=o},
    b3c/.style={thick, color=\bthree, mark=o},
    b4c/.style={thick, color=\bfour, mark=o},
    crcr/.style={thick, color=LimeGreen, mark=o},
    crcrp/.style={thick, color=LimeGreen, mark=o,mark size=8pt},
    crca/.style={thick, color=OliveGreen, mark=diamond},
    crcap/.style={thick, color=OliveGreen, mark=diamond,mark size=8pt},
    mae/.style={thick, color=Gray, mark=star},
    maep/.style={thick, color=Gray, mark=star, mark size=8pt},
    y1c/.style={thick, color=Goldenrod, mark=o},
    r1c/.style={thick, color=Maroon, mark=o},
    r2c/.style={thick, color=WildStrawberry, mark=o},
    r3c/.style={thick, color=RedViolet, mark=o},
    y1c/.style={thick, color=Goldenrod, mark=o},
    y2c/.style={thick, color=Apricot, mark=o},
    y3c/.style={thick, color=BurntOrange, mark=o},
    y3cp/.style={thick, color=BurntOrange, mark=o, mark size=8pt},
    yvert/.style={dashed, thick, color=BurntOrange, mark=o},
    yvertp/.style={dashed, thick, color=BurntOrange, mark=o, mark size=8pt},
    tt3/.style={thick, color=Maroon, mark=o},
    tt4/.style={thick, color=BurntOrange, mark=o},
    tt1/.style={thick, color=Maroon, mark=star},
    tt2/.style={thick, color=BurntOrange, mark=star},
    ttl1/.style={thick, color=Maroon},
    ttl2/.style={thick, color=BurntOrange},
    ttl3/.style={thick, mark=star,mark size=3pt},
    ttl4/.style={thick, mark=o},
}

%% file: plots/seg_f_mask.tex
\begin{tikzpicture}
\begin{axis}[%
  width=1.3\linewidth,
  height=4cm, 
   xlabel={\footnotesize Masking ratio},
   xlabel style={yshift=4pt},
  minor tick num=1,
  ymin=34.5, ymax=39.5,
  ticklabel style = {font=\scriptsize},
  legend pos=north east,
    title={\footnotesize ADE Semantic segm. (mIoU) $\uparrow$},
            title style={yshift=-8pt},
  ]

\input{plots/ffs3d_table}
    \addplot[y3c]      table[x=rgb_mask,  y=seg_mask]   \map; 
    \addplot[yvert] coordinates {(0.9, -1) (0.9, 50)};  
   \draw[fill,BurntOrange] (axis cs: 0.9,38.96) circle (2pt);

\end{axis}
\end{tikzpicture}

%% file: plots/ffs3d_table.tex
\pgfplotstableread{
perf_rgb_mask   rgb_mask   task_avg_mask  seg_mask  task_rank_mask
80.2 		0.75       37.3133        35.14     6.8
79.2 		0.8        37.4255        35.43     5.8
81.4 		0.85       35.9190        36.98     5.0
83.4           0.9         34.7359        38.96       2.6
82.0           0.95        35.5497        36.36     3.8

}{\map}

\pgfplotstableread{
de_cross_grad  	de_cat_grad	de_mae  iter   tasko_avg_cross_grad 	tasko_avg_cat_grad      tasko_avg_mae   seg_cat_grad    seg_cross_grad  seg_mae  tasko_rank_cat  tasko_rank_grad tasko_rank_mae
68.4             73.9		70.0	20     41.9912   	    	41.8576			40.2027		32.16	        31.93           34.52    7.166           6.33            6.125
78.5             81.1		75.0	40     37.7156   	    	37.3443			37.2879		34.94	        35.16           36.80      6.042           4.5           6.125
81.3             84.1		79.0	60     35.4377   	    	35.5734			36.5511		36.51	        35.99           37.97    4.375           3.0             5.5
82.7             85.4		79.0	80     35.1754   	    	35.1979			36.8924		37.24	        37.12           38.22    3.75            2.5             5.25
84.8             86.5		79.0	100    34.6570   	    	34.5193			36.6434		38.84	        38.26           38.63    3.875           3.0             5.54
86.8             87.4		79.0	200    33.5619   	    	34.0330			36.5866		39.98	        38.75           38.94    1.70            2.375           4.75
86.1             86.2		79.0	400    33.0035   	    	33.3526			35.6532		41.30	        40.63           40.26    2.125           1.08            4.3
}{\map}

%% file: plots/depth_f_mask.tex
\begin{tikzpicture}
\begin{axis}[%
  width=1.3\linewidth,
  height=4cm, 
   xlabel={\footnotesize Masking ratio},
   xlabel style={yshift=4pt},
  minor tick num=1,
  ticklabel style = {font=\scriptsize},
  ymin=78, ymax=85,
  legend pos=north east,
    title={\footnotesize NYUv2 depth (Acc@1.25) $\uparrow$},
            title style={yshift=-8pt},
  ]

\input{plots/ffs3d_table}
    \addplot[y3c]      table[x=rgb_mask,  y=perf_rgb_mask]   \map; 
    \draw[fill,BurntOrange] (axis cs: 0.9,83.4 ) circle (2pt);
     \addplot[yvert] coordinates {(0.9, -1) (0.9, 200)};  


\end{axis}
\end{tikzpicture}

%% file: plots/taskonomy_f_mask.tex
\begin{tikzpicture}
\begin{axis}[%
  width=1.3\linewidth,
  height=4cm, 
   xlabel={\footnotesize Masking ratio},
   xlabel style={yshift=4pt},
  ticklabel style = {font=\scriptsize},
  minor tick num=1,
  ymin=32.5, ymax=39,
  legend pos=north east,
    title={\footnotesize Taskonomy (avg) $\downarrow$},
            title style={yshift=-8pt},
  ]

\input{plots/ffs3d_table}
    \addplot[y3c]      table[x=rgb_mask,  y=task_avg_mask]   \map; 
             \draw[fill,BurntOrange] (axis cs: 0.9,34.7359) circle (2pt);
             \addplot[yvert] coordinates {(0.9, -1) (0.9, 50)};  
   
\end{axis}
\end{tikzpicture}

%% file: plots/legend_f_epoch.tex
\begin{center}
\resizebox{.6\linewidth}{!}
{
    \begin{tikzpicture}
    \begin{axis}[%
        hide axis,
        xmin=10, xmax=50,
        ymin=0,ymax=0.4,
        legend columns=-1,
        legend style={draw=white!15!black,legend cell align=left}
        ]
        \addlegendimage{crcr}
        \addlegendentry{CroCo (CrossBlock)}
        \addlegendimage{crca} 
        \addlegendentry{CroCo (CatBlock)}
        \addlegendimage{mae}
        \addlegendentry{MAE (habitat)}
    \end{axis}
    \end{tikzpicture}
    }
    \end{center}

%% file: plots/seg_f_epoch.tex
\begin{tikzpicture}
\begin{axis}[%
  width=1.2\linewidth,
  height=4.5cm, 
   xlabel={\footnotesize Pre-training epochs},
   xlabel style={yshift=4pt},
  minor tick num=1,
  ticklabel style = {font=\scriptsize},
  legend pos=north east,
    title={\footnotesize ADE Semantic segm. (mIoU)  $\uparrow$},
            title style={yshift=-8pt},
  ]

\input{plots/ffs3d_table}
    \addplot[crcr]      table[x=iter,  y=seg_cross_grad]   \map; 
    \addplot[crca]      table[x=iter,  y=seg_cat_grad]   \map; 
    \addplot[mae]      table[x=iter,  y=seg_mae]   
    \map; 
   
\end{axis}
\end{tikzpicture}

%% file: plots/depth_f_epoch.tex
\begin{tikzpicture}
\begin{axis}[%
  width=1.2\linewidth,
  height=4.5cm, 
   xlabel={\footnotesize Pre-training epochs},
   xlabel style={yshift=4pt},
  minor tick num=1,
  ticklabel style = {font=\scriptsize},
  legend pos=north east,
    title={\footnotesize NYUv2 depth (Acc@1.25) $\uparrow$},
            title style={yshift=-8pt},
  ]

\input{plots/ffs3d_table}
    \addplot[crcr]      table[x=iter,  y=de_cross_grad] \map;
    \addplot[crca]      table[x=iter,  y=de_cat_grad] \map;
    \addplot[mae]      table[x=iter,  y=de_mae]  \map; 
   
\end{axis}
\end{tikzpicture}

%% file: plots/taskonomy_f_epoch.tex
\begin{tikzpicture}
\begin{axis}[%
  width=1.2\linewidth,
  height=4.5cm, 
   xlabel={\footnotesize Pre-training epochs},
    xlabel style={yshift=4pt},
  minor tick num=1,
  legend pos=north east,
    ticklabel style = {font=\scriptsize},
    title={\footnotesize Taskonomy (avg) $\downarrow$},
            title style={yshift=-8pt,xshift=10pt},
  ]
 	      
\input{plots/ffs3d_table}
    \addplot[crcr]      table[x=iter,  y=tasko_avg_cross_grad]   \map; 
    \addplot[crca]      table[x=iter,  y=tasko_avg_cat_grad]   \map; 
    \addplot[mae]      table[x=iter,  y=tasko_avg_mae]   \map; 
   
\end{axis}
\end{tikzpicture}

%% file: plots/stereo-vkitti.tex
\begin{table}[h]
\centering
\small
\begin{tabular}{lccccccccccr}
\toprule
Method/     & fog& sun-& clone& over-& rain& mor-& 15$^{\circ}$-& 15$^{\circ}$-& 30$^{\circ}$-& 30$^{\circ}$- & Ave-\\
Pre-training &	 & set &      & cast &     & ning& left         & right        & left         & right  &rage \\
\midrule
MAE (Habitat) 
      &	1.80& 1.75& 1.96& 1.89&	2.13& 2.17& 2.93& 2.07& 4.06& 2.25& 2.30 \\
{\color{OliveGreen}\bf{CroCo}} (Habitat)
      & \bf{1.15}& 1.69& 1.72& 1.49& 1.77&\underline{1.68}& 2.49&\underline{1.58}& 3.43& \underline{1.78}& 1.88 \\ 
\midrule
PSMNet~\cite{chang2018pyramid} 
      & 2.36 & 2.30 & 2.38 & 2.45 & 2.31 & 2.39 & 2.36 & 2.41 & 2.33 & 2.25 &2.36\\
LaC-GwcN~\cite{liu22localsimilarity} 
      & 1.67 & \bf{1.41} &\underline{1.52}&\underline{1.39}&\underline{1.30}&1.72 &\underline{1.65}&\bf{1.51}& \underline{1.50}& 1.79&\underline{1.54}\\
LEAStereo~\cite{cheng20leastereo} 
      &\underline{1.24}&\underline{1.42}&\bf{1.27}&\bf{1.17}&\bf{1.05}&\bf{1.09}&\bf{1.18} &1.65 &\bf{1.06} & \bf{1.22}&\bf{1.23} \\
\bottomrule
\end{tabular}
\caption{\textbf{Stereo matching results} with the average 3-px error for 10 VKITTI variants. We compare CroCo pre-training with MAE pre-training as well as other state-of-the-art methods. Best result per column on bold and second best underlined.}
\label{tab:stereo-vkitti}
\end{table}